%% file: main.tex
\documentclass[twocolumn]{article}

\usepackage{framed,multirow}
\usepackage{amssymb}
\usepackage{latexsym}
\usepackage{url}
\usepackage{xcolor}
\usepackage{graphicx}

\usepackage{hyperref}
\usepackage{booktabs}
\usepackage{placeins}
\usepackage{pbox}
\usepackage{amsmath}
\usepackage{multirow}
\usepackage{float} 
\usepackage[bottom=1.8cm, right=1.3cm, left=1.3cm, top=1.3cm]{geometry}

\title{Unsupervised pre-training of graph transformers on patient population graphs}

\author{Chantal Pellegrini$^1$\thanks{Corresponding author: chantal.pellegrini@tum.de. One of the datasets used in preparation of this article was obtained from the Alzheimer’s Disease Neuroimaging Initiative (ADNI) database (adni.loni.usc.edu). As such, the investigators within the ADNI contributed to the design and implementation of ADNI and/or provided data but did not participate in analysis or writing of this report. A complete listing of ADNI investigators can be found at: \url{http://adni.loni.usc.edu/wp-content/uploads/how_to_apply/ADNI_Acknowledgement_List.pdf}} \and Nassir Navab$^{1,2}$ \and Anees Kazi$^{1,3}$\thanks{The work was done while A. Kazi was affiliated with the TU Munich.}}
\date{%
    $^1$Computer Aided Medical Procedures, Technical University Munich, Germany\\%
    $^2$Computer Aided Medical Procedures, Johns Hopkins University, Baltimore, USA\\%
    $^3$Massachusetts General Hospital, Harvard Medical School, Cambridge, Massachusetts, USA\\[2ex]%
}

\begin{document}
\maketitle

\begin{abstract}
\textbf{Pre-training has shown success in different areas of machine learning, such as Computer Vision, Natural Language Processing (NLP), and medical imaging. However, it has not been fully explored for clinical data analysis. An immense amount of clinical records are recorded, but still, data and labels can be scarce for data collected in small hospitals or dealing with rare diseases. In such scenarios, pre-training on a larger set of unlabelled clinical data could improve performance. In this paper, we propose novel unsupervised pre-training techniques designed for heterogeneous, multi-modal clinical data for patient outcome prediction inspired by masked language modeling (MLM), by leveraging graph deep learning over population graphs. To this end, we further propose a graph-transformer-based network, designed to handle heterogeneous clinical data. By combining masking-based pre-training with a transformer-based network, we translate the success of masking-based pre-training in other domains to heterogeneous clinical data. We show the benefit of our pre-training method in a self-supervised and a transfer learning setting, utilizing three medical datasets TADPOLE, MIMIC-III, and a Sepsis Prediction Dataset. We find that our proposed pre-training methods help in modeling the data at a patient and population level and improve performance in different fine-tuning tasks on all datasets.}
\end{abstract}

\input{chapters/introduction}
\input{chapters/method}
\input{chapters/experiments_results}
\input{chapters/conclusion}

\bibliographystyle{ieeetr}
\bibliography{refs} 

\end{document}

%% file: chapters/introduction.tex
\section{Introduction}

A large amount of medical data is collected on a daily basis in many different hospitals. Often this data is stored in Electronic Health Records (EHRs), digital patient charts containing various information such as the patient's medical history, diagnoses, treatments, medical images, and lab or test results. However, even though a large amount of data is recorded, for some tasks labeled data is scarce, as labeling can be tedious, time-consuming, and expensive (\cite{xiao2018opportunities}). Nevertheless, their digital and structured form can enable easy access to apply learning-based methods to EHR data, in particular unsupervised methods (\cite{landi2020deep}). Further, data collected in small hospitals or for rare diseases is often limited (\cite{mitani2020small}). In these scenarios, the ability to leverage the large body of unlabeled clinical data could boost the performance and confidence of prediction systems on these smaller datasets. This, in turn, can support clinicians in better and faster diagnosis and decision-making.

Clinical records contain multiple heterogeneous data types, including imaging and non-imaging data. While non-imaging data usually is in numerical format, medical images consist of 2D or 3D data. In this work, we use imaging biomarkers to convert the imaging data to numerical features allowing a simple fusion of multi-modal input and reducing the computational requirements. This allows us a very general formulation of our method, which can be applied to both static and longitudinal data, handling both imaging and non-imaging features simultaneously. In our ablation studies, we further show that our model can be extended to deal with spatial images.

Unsupervised pre-training can be a useful tool to exploit unlabeled data and has shown great success in other domains, such as natural language processing (NLP) (\cite{word2vec,pennington2014glove,elmo,radford2018improving,devlin2018bert,NEURIPS2019_dc6a7e65,liu2019roberta}), computer vision (CV) (\cite{pathak2016context,bojanowski2017unsupervised,caron2018deep,komodakis2018unsupervised,bao2021beit}) and medical imaging (\cite{bai2019self,chen2019self,ouyang2020self}). In these domains various types of pre-training have been proposed, including generative approaches e.g. based on auto-encoders, contrastive learning, and the application of hand-crafted pre-text tasks such as masked content prediction (\cite{liu2021self,han2021pre}). In the medical domain, next to medical imaging, pre-training was applied for instance on medical code data (\cite{shang2019pre,li2020behrt,rasmy2021medbert,pang2021cehr}) and textual EHR data (\cite{med_rep_learning_park}). However, it is not explored enough for complex clinical data such as heterogeneous, multi-modal patient records (EHRs). Only a few previous works investigated pre-training for this type of data, laying the ground-stone for pre-training on heterogeneous EHRs (\cite{mcdermott2021EHRbenchmark,gupta2020transfer}). They show the benefit of pre-training, especially for scenarios with limited labeled data. Nevertheless, their improvements via pre-training are limited, in particular, if more labeled data is available, and their pre-training task designs do not fully take the longitudinal and complementary nature of clinical records into account.

In the papers mentioned above the community explored various model architectures, including RNNs, GRUs, and Transformers. For sequential data, such as natural language and time-series data, transformer models (\cite{vaswani2017attention}) are currently dominant. These models often are combined with self-supervised masking-based pre-training tasks, which has proven to be a promising combination (\cite{devlin2018bert,bao2021beit,li2020behrt}). In this work, we focus on this type of self-supervised pre-training and combine it with a graph-transformer-based network.

In the recent literature, a new way for multi-modal patient data analysis using patient population graphs is getting explored. In a population graph, each node $n_i$ in the graph $G$ represents a patient and the edges in G incorporate the similarities between the patients. Such patient population graphs have been leveraged to help analyze patients' medical data using the relationship among different patients (\cite{survey_medical_graph_dl}). This enables clinically semantic modeling of the data. By using the patient population graph during unsupervised pre-training, node representations based on feature similarities between the subjects can be learned. These representations can improve the understanding of the data, which can then help to improve patient-level predictions. Therefore we model the EHR data in a patient population graph for both pre-training and fine-tuning.

Several works successfully apply pre-training to graph data and show the benefits on both common graph benchmarks (\cite{zhang2020graphbert,hu2020gpt}) as well as domain-specific data such as molecular or biological graphs (\cite{hu2019strategies,rong2020grover,lu2021learning}). However, these methods are often specialized to a certain domain, such as protein or molecule graphs, or focus on improving the graph-level embedding. To the best of our knowledge, pre-training was not previously applied to patient population graphs, where the pre-training task needs to be formulated to foster meaningful node-level embeddings, representing the patient's data. As patient population graphs have proven useful for outcome and disease prediction tasks on clinical data, we believe a well-working pre-training technique for patient population graphs is of high value to the community and has the potential to improve performance in many patient-level prediction tasks.

In this paper, we propose a graph transformer-based model suitable for learning on multi-modal clinical data in form of population graphs. Motivated by the huge success of masked language modeling pre-training for transformer models, we propose masking-based pre-training methods and combine them with our graph transformer-based architecture. Our pre-training methods are specialized for multimodal clinical data. Our main contributions are:
\begin{itemize}
   \item We develop multiple novel unsupervised pre-training methods based on masked imputation of clinical input features, which are specifically designed for (longitudinal) EHRs and multi-modal clinical data modeled as population graphs.
   \item We propose a novel (graph) transformer-based model suitable to learn over heterogeneous clinical data allowing multi-modal data fusion. The intelligent design and combination of state-of-the-art building blocks allow us to deal with heterogeneous data and show the potential of transformers for multi-modal clinical data analysis. Our model is designed to handle various input data types occurring in multi-modal clinical records, taking static and time-series data and continuous as well as discrete numerical features into account.
   \item By combining our graph transformer-based model with masking-based pre-training, we show a way to translate the success of pre-training of transformers in other domains to multi-modal clinical data.
   \item We show significant performance gains through pre-training when fine-tuning with as little as 1\% and up to 100\% labels in both the self-supervised and the transfer learning setup, providing a solution to limited labeled data.
\end{itemize}

We evaluate our method over general EHR data as well as brain imaging data over two publicly available, medical datasets, MIMIC-III (\cite{mimic}) and TADPOLE (\cite{tadpole}) for Length-of-Stay prediction (\cite{zebin2019deep,wang2020mimicextract,mcdermott2021EHRbenchmark}), combined Discharge and Mortality Prediction (\cite{mcdermott2021EHRbenchmark}) and Alzheimer's disease prediction (\cite{parisot2018disease,kazi2019self,cosmo2020latent}). We provide an extensive analysis of our model and the effects of the different pre-training approaches on fine-tuning performance for different amounts of labels. Further, we test our method in a transfer learning setting, where we pre-train on the MIMIC-III dataset and fine-tune on the Sepsis Prediction dataset published in the PhysioNet/Computing in Cardiology Challenge 2019 (\cite{reyna2019early_sepsis_ds,goldberger2000physiobank}). Our source code is available at \href{https://github.com/ChantalMP/Unsupervised_pre-training_of_graph_transformers_on_patient_population_graphs}{\url{https://github.com/ChantalMP/Unsupervised_pre-training_of_graph_transformers_on_patient_population_graphs}}.

The remainder of this paper is structured as follows. After presenting the most relevant related work in Section \ref{rel_work}, we provide a detailed description of our method in Section \ref{method}, including population graph construction, the proposed model architecture, and the proposed pre-training strategies. In Section \ref{experiments} we first introduce the datasets we use and our experimental setup and then show our results, as well as their interpretation and discussion. Finally, a conclusion is given in Section \ref{conclusion}.

\section{Related Work} \label{rel_work}
The main contribution of our method lies in investigating unsupervised pre-training for heterogeneous EHRs modeled as population graphs for patient-outcome prediction. Accordingly, we divided this section into multiple parts focusing on patient population graphs and pre-training for graph and EHR data.

\subsection{Population Graphs for Patient Outcome Prediction}
Several previous works use patient population graphs in combination with graph neural networks in the field of diagnosis and patient outcome prediction (\cite{survey_medical_graph_dl}). \cite{parisot2018disease} introduced the use of GCNs for the analysis of population graphs in the medical domain for combining imaging and non-imaging data. Given a population of patients, they construct a fully connected graph, where the edges are based on the similarity of non-imaging data, while the node features describe a patient's imaging data. InceptionGCN (\cite{kazi2019inceptiongcn}) introduces an inception module for spectral GCNs, showing one way to deal with heterogeneous graphs to model multi-modal data. \cite{valenchon} propose to use multiple graphs based on different subject features. In another work, \cite{kazi2019self} propose to use an attention mechanism for weighting the subject's demographic features. Several works aim to improve the graph structure by either updating a pre-constructed graph during training (\cite{huang2020edge}) or learning the optimal graph end-to-end (\cite{cosmo2020latent,kazi2022differentiable}). Other works focus on dealing with missing (\cite{vivar2018multi}) or imbalanced data (\cite{ghorbani2022ra}). Overall, modeling clinical data in a population graph has been proven a promising direction in patient outcome prediction.

\subsection{Masking-based Pre-training}
In the NLP domain, pre-training was adopted mainly using self-supervised learning, aimed to understand the intrinsics of natural language without the need for any human supervision (\cite{qiu2020pre}). This allows to exploit the huge corpora of unlabeled text data. BERT (\cite{devlin2018bert}) is one of the most important works about pre-training transformer-based language models. It is designed to learn bidirectional representations from unlabeled text. After being fine-tuned with task-specific data, it has set a new state-of-the-art in many NLP tasks. They introduce masked language modeling (MLM), which until now is one of the most used pre-training tasks for NLP and was further adapted to be used in several other domains (\cite{yu2021point,bao2021beit,li2020behrt}). Given an input text, during MLM, 15\% of the input tokens are randomly masked and replaced with a mask token '[MASK]'. After being processed by several transformer layers, the final hidden representations are fed into an output softmax over the defined vocabulary aiming to predict the masked token. For fine-tuning, the model is initialized with the pre-trained weights, only the final prediction layers are changed according to the task. In the medical domain, BERT-like pre-training was applied for tasks such as disease prediction (\cite{rasmy2021medbert}), medication recommendation (\cite{shang2019pre}), medical imaging \cite{wang2021transpath}, clinical outcome prediction (\cite{pang2021cehr,mcdermott2021EHRbenchmark}) and clinical NLP tasks (\cite{alsentzer2019publicly,he2020infusing}). However, it was only rudimentary explored for heterogeneous EHR data.

\subsection{Pre-Training on Graph Data}
The previously proposed pre-training techniques for graph data reach from node-level to graph-level to generative tasks. These tasks include tasks like attribute reconstruction, graph-level property prediction, or auto-regressive generation of nodes and edges (\cite{xie2022self}). \cite{hu2019strategies} propose to combine node and graph level pre-training by first using a node-level task such as attribute masking to train the model, followed by training to predict graph level properties (e.g. properties of a certain protein). \cite{rong2020grover} propose a transformer-based network, pre-trained by predicting properties of masked sub-graphs (such as the number of neighbors of a certain atom type) and the presence of certain motifs (recurrent sub-graphs such as functional groups in molecules) in a graph. \cite{zhang2020graphbert} propose Graph-BERT, a translation of BERT pre-training to the graph domain. Instead of processing a full graph, they create link-less sub-graphs of a node's neighborhood, add certain positional embeddings to each node, and process them by a conventional transformer. As pre-training tasks, they propose to reconstruct a node's attributes using a linear layer and to optimize the model to have high cosine-similarities between the embeddings of nodes with a high ground truth intimacy. \cite{hu2020gpt} propose a generative graph pre-training framework called GPT-GNN. For pre-training, both attributes and edges are generated, given a partial graph as initialization. While the previous works focus on unsupervised pre-training within one graph domain, \cite{verma2019learning} investigated transfer learning for graph data. Their model consists of an input transformer, a simple linear layer, a graph encoder, which is a conventional GNN, and a task-specific graph decoder. While the input transformer and the graph decoder are dataset-specific, the graph encoder is trained jointly on several graph datasets from the bioinformatics and social network domain. They improve classification results in comparison to classical GNNs and graph kernel methods. The aforementioned methods could show the applicability of pre-training in the graph domain. However, to the best of our knowledge, no previous work investigated pre-training for patient population graphs.

\subsection{Pre-Training on EHR data}
The most general form of multi-modal clinical records are Electronic Health Records, which can contain any data recorded over a patient's stay in a hospital. As EHRs are usually recorded throughout the patient's stay, they often have a sequential structure, making sequence models like transformers a good fit to learn on this data. Some previous works study how to pre-train BERT (\cite{devlin2018bert}) over simplified EHR records for downstream tasks in disease and medication code prediction. Here the input records contain only sequences of medical codes, such as diagnosis or medication codes but do not include multi-modal heterogeneous data. In many of these works, an adapted version of MLM is used as a pre-training task (\cite{shang2019pre,li2020behrt,rasmy2021medbert,pang2021cehr}). \cite{agrawal2022leveraging} investigate order-contrastive pre-training for clinical time-series data and test their method on synthetic data and temporally ordered clinical radiology notes. \cite{mcdermott2021EHRbenchmark} propose one pre-training approach over heterogeneous EHR data. They create a pre-training benchmark over the eICU \cite{pollard2018eicu} and MIMIC-III \cite{mimic} datasets and come up with two baseline pre-training methods, unsupervised masked imputation, where random time points are masked, and supervised multi-task learning. Furthermore, they define several downstream tasks to evaluate their method. Pre-training and fine-tuning are performed over records of single patient stays in the ICU using a bi-directional Gated Recurrent Unit. \cite{gupta2020transfer} present an approach for transfer learning to improve outcome prediction on longitudinal EHR records by processing each input feature separately with a pre-trained TimeNet \cite{malhotra2017timenet}, the encoder of an auto-encoder RNN, which was pre-trained to reconstruct diverse univariate time-series data. The aforementioned methods have one common drawback, as they often simplify the given EHR data in order to apply pre-training. The few methods working with heterogeneous EHRs, do still not fully consider the form of EHR data in their pre-training task design and show only limited improvements.
\medskip

For the graph domain, several works show the promise of pre-training. However, no work investigated pre-training for patient population graphs, where it can lead to better patient outcome predictions. On the other hand, work on pre-training for EHR data remains very limited. The existing previous methods do not fully account for their longitudinal and heterogeneous nature and show limited improvements. We improve upon these works and propose novel EHR-specific pre-training techniques based on masking for EHR data modeled as population graphs. In the next section, we explain our method in detail.

%% file: chapters/method.tex
\section{Method} \label{method}
Our method is designed to target patient-level prediction tasks on a dataset $\mathbf{D}$, composed of the clinical records of \textit{N} patients. Toward this, we propose a two-step pipeline. The first step entails the use of unsupervised pre-training methods to enhance the general understanding of the data by our model. In the second step, we fine-tune the pre-trained model on a downstream prediction task $T$. The understanding learned via unsupervised pre-training aims to improve the performance on downstream tasks despite limited labeled data.

In the dataset $\mathbf{D}$, let $n_i$ and $r_i$ be the ith patient and its corresponding record, where $i \in [1,N]$. Every record $\mathbf{r_i}$ consists of a set of heterogeneous input features, denoted by $\mathbf{r_i} \subseteq [\mathbf{d_i},\mathbf{c_i},\mathbf{t_{d_i}},\mathbf{t_{c_i}}]$. Each record can contain all or a subset of these features. The feature types include static discrete features $\mathbf{d_i} \in \mathbb{N}^D$, static continuous features $\mathbf{c_i} \in \mathbb{R}^C$, and discrete as well as continuous time-series features  $\mathbf{t_{d_i}} \in \mathbb{R}^{S_d \times \tau }$ and $\mathbf{t_{c_i}} \in \mathbb{R}^{S_c \times \tau }$, where $S_{d/c}$ denotes the feature dimension and $\tau $ the length of the time-series. Static features $\mathbf{d_i}$ and $\mathbf{c_i}$ encompass different numerical values such as patient demographics or measurements taken once for every patient, whereas the time-series features $\mathbf{t_{d_i}}$ and $\mathbf{t_{c_i}}$ consist of a sequence of values recorded throughout a patient's stay in the hospital, such as vital signs (e.g. heart rate, temperature) or repeatedly measured lab values (e.g. Hemoglobin, glucose level). All targeted downstream tasks $T$ are binary or multi-class classification tasks on a patient level with labels $\mathbf{Y} \in \mathbb{N}^L$ given for L classes, and the task is to make predictions for all patient records in the test set.

As the same model is used for pre-training and fine-tuning, to understand our pipeline we will start by introducing the proposed model architecture including the different sub-modules and the population graph construction. Afterward, we give details about the pre-training and fine-tuning procedure.

\subsection{Model Architecture}
Our model is tasked with making node-level predictions for every node in the graph $\mathit{G}$, based on the input features $\mathbf{d},\mathbf{c},\mathbf{t_{d}}$ and $\mathbf{t_{c}}$. For this task, we propose a model, consisting of an encoder and a decoder. The encoder is used to generate meaningful node representations given the input EHR graph. Thereafter the decoder's task is to capture the essence of features inclined toward a node-level classification task. The same model is used in the pre-training as well as the fine-tuning stage.

\begin{figure*}[hbt!]
\centering
\includegraphics[width=\textwidth]{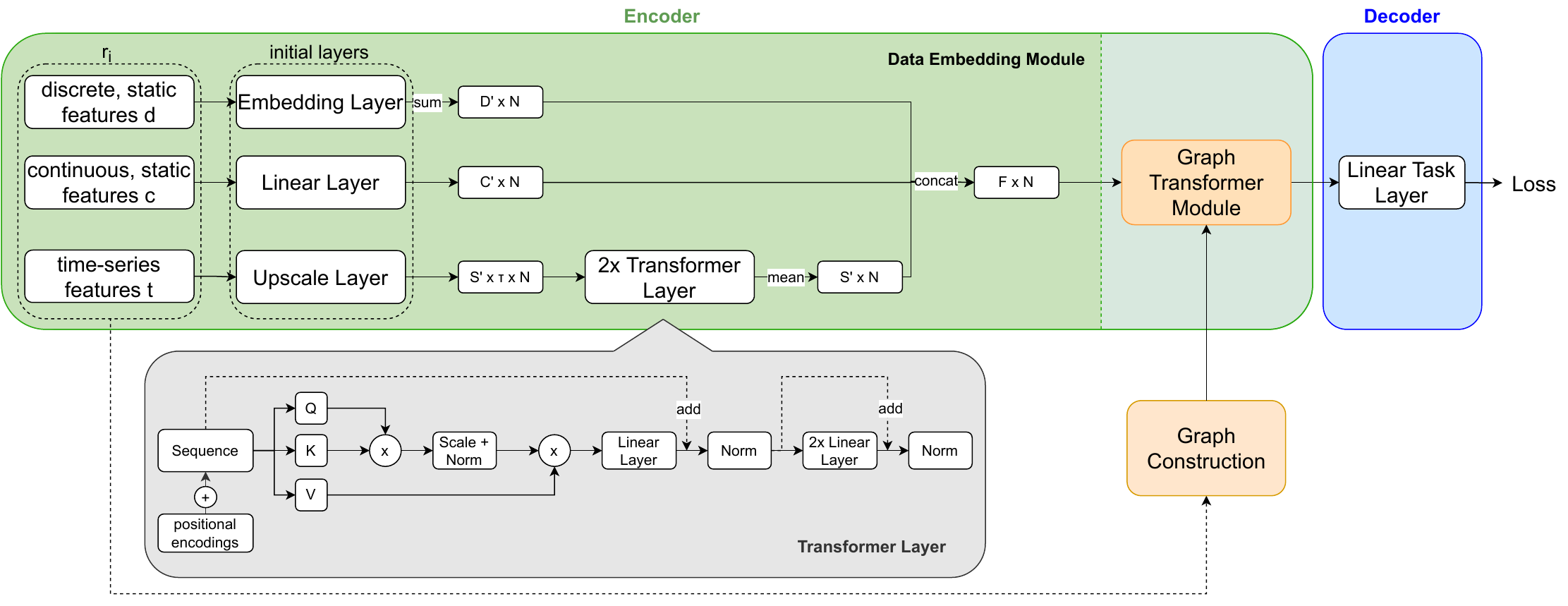}
\caption{Overview of the proposed architecture. All input features are combined into one node embedding, applying transformer layers to enhance the time-series features. The resulting graph is processed by several Graphormer layers and a linear task layer. The population graph is constructed using the original input features $r_i$.}
\label{arch_overview}
\end{figure*}
\subsubsection{Encoder}
The encoder comprises a data embedding module and a graph transformer module, based upon Graphormer \cite{graphormer}. The data embedding module is responsible to fuse the heterogeneous input features occurring in multi-modal clinical records. The fused features are then used as node embeddings in our constructed population graph which is processed by the graph transformer module.

\textbf{Data embedding module:} 

The data embedding module converts the heterogeneous input features, $\mathbf{d},\mathbf{c},\mathbf{t_{d}}$ and $\mathbf{t_{c}}$, into one common representation to form node embeddings for every node in the graph. 

To process static, discrete input features, we follow the conventional Graphormer (\cite{graphormer}) and apply an embedding layer, followed by a summation over the feature dimension. This transforms the input features $\mathbf{d_i}$ into embedded features $\mathbf{d'_i} \in \mathbb{R}^{D'}$. 

While Graphormer is limited to static, discrete input features only, we extend the model to support static, continuous input features as well. As for continuous features embedding layers are not applicable, these features are processed by a linear layer. Again this results in an embedding vector $\mathbf{c'_i} \in \mathbb{R}^{C'}$. 

Next to static features, we also support discrete and continuous time-series features $\mathbf{t_{{d/c}_i}} \in \mathbb{R}^{S_{d/c} \times \tau}$. We first upscale the feature dimension of every time step by applying an upscale layer $U_\theta$ with weights $\theta$. For discrete features, this upscale layer is an embedding layer and for continuous features, a linear layer, followed by a summation over the feature dimension. These up-scaled features $\mathbf{t^{(u)}_{{d/c}_i}}$ are further processed by two transformer layers, as described by \cite{vaswani2017attention}, to enhance the embeddings using temporal context. The transformer layers output adapted embeddings $(e_1, e_2, ..., e_\tau)$ per time-step. The final embedding for the time-series features $\mathbf{t'_{{d/c}_i}}$, is then formed as the mean of the time-step embeddings $(e_1, e_2, ..., e_\tau)$, which allows working with sequences of variable lengths. This results in embedded features $\mathbf{t'_{d_i}} \in \mathbb{R}^{S'_d}$ for the discrete and $\mathbf{t'_{c_i}} \in \mathbb{R}^{S'_c}$ for continuous features.

The feature vectors $\mathbf{d'_i}, \mathbf{c'_i}$, $\mathbf{t'_{d_i}}$ and $\mathbf{t'_{c_i}}$ can now be concatenated to form the final node embeddings $n_i \in \mathbb{R}^{F}$  for each of the \textit{N} nodes, where $F = \sum_{F_k \subseteq [D', C', S'_d, S'_c]} F_k$ is the sum of the feature dimensions of the transformed input features.

\begin{figure*}[hbt!]
\centering
\includegraphics[width=\textwidth]{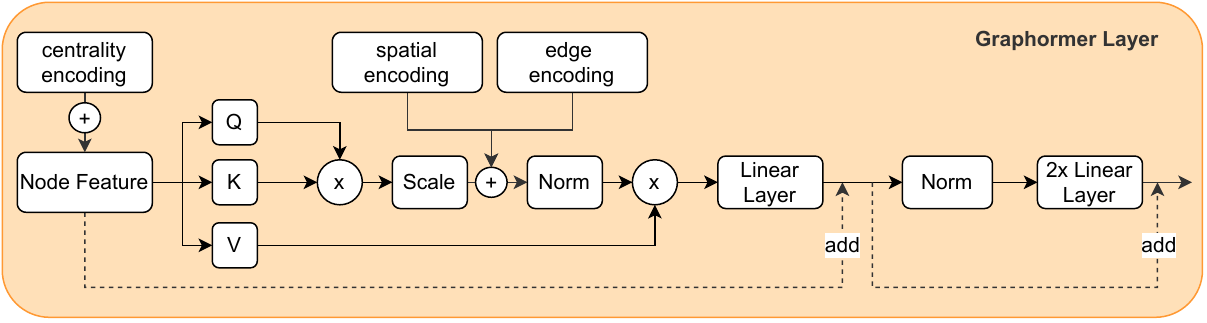}
\caption{Overview of a Graphormer layer as proposed by\cite{graphormer}. Graphormer is a graph transformer model based on global attention to all nodes. The graph structure is integrated via different structural encodings. The centrality encoding, which encodes the in and out degree of each node is added directly to the node features. Given the shortest path between two nodes, the spatial and edge encoding encode the length and the edge features of the path. They are both added as a bias to the attention.}
\label{graphormer_layer}
\end{figure*}

\textbf{Graphormer Module:} The backbone of our model comprises $\mathbf{L}$ graph transformer layers as proposed by \cite{graphormer}. Dependent on the dataset we vary the number of Graphormer layers. Graphormer is based upon transformer (\cite{vaswani2017attention}) and outperforms traditional graph neural networks in several graph-level prediction tasks. Their main contribution is the encoding of the graph structure, without restricting attention to neighborhoods. Instead, they attend to all nodes in the graph and incorporate the graph structure via structural encodings, which encode the in and out degrees of the nodes, the length of the shortest path between each node pair, and the edge features lying on this path. The encoding of the in and out degree is added directly to the node features of every node, while the other structural encodings are added as a bias to the attention score between two nodes. Figure \ref{graphormer_layer} shows an overview of a Graphormer layer.\\
The input graph for Graphormer is constructed based on the features of the input records $r_i$ as described in the following paragraph.

\textbf{Graph Construction:}
Let $\mathit{G} = (V,E)$ be a population graph where the nodes $V = \left \{n_1,...,n_N \right \}$ represent the patients. Each node is associated with features coming from the patient record $r_i$, and the edges $E = \left \{e_{i,j}: i,j \in N\right \}$ define the connections between the nodes. For each node pair $n_i$ and $n_j$ representing the records $r_i$ and $r_j$, we calculate a similarity score $S(r_i, r_j)$ between the features to decide if there exists an edge $e_{i,j}$. As we have various feature types, we define different similarity scores, which are suitable for the respective features, based on L2 distance for continuous features and absolute matching for discrete features (\cite{survey_medical_graph_dl}). After computing the similarities for all feature types we average them to get one overall similarity score for each record pair. Before averaging, all type-specific similarity scores are normalized to a range between zero and one using min-max normalization or sigmoid. As our focus does not lie on graph construction, we choose a conventional graph construction method and construct a k-NN graph with $k=5$. The value of $k$ is chosen experimentally to avoid many disconnected components as well as very densely connected regions (see supplementary material). We want to avoid disconnected as well as very densely connected regions, as in both extremes the graph structure does not give any valuable information about the relationships between the nodes. This similarity score is additionally used as the edge feature of the edge between two nodes. This enables the model to give more attention to nodes connected over a path with high feature similarities. The experiment section contains a detailed description of the graph construction per dataset.

\subsubsection{Decoder}
The decoder consists of one or multiple linear layers for predicting outputs for the current pre-training or downstream task. Depending on the task, the decoder layer is adapted to fit the desired output dimensions. When training for an end-task the decoder is a linear output layer, with an output dimension according to the number of classes in the targeted task. During pre-training, the decoder also consists of linear layers, one for each input data type. This layer predicts an output vector in the same shape as the input data, where each element can be interpreted as the prediction for the corresponding input value. Even though we make a prediction for all input values, only the predictions of masked values are used to update the model.

\subsection{Training Pipeline} 
Until now we described the model architecture needed for fine-tuning and pre-training and its components. Now we will explain the two main steps of our pipeline, pre-training and fine-tuning, in detail.

\subsubsection{Pre-training Step}

We propose several unsupervised pre-training strategies for patient population graphs of Electronic Health Records. All proposed pre-training methods use the paradigm introduced with masked language modeling and task the model to predict masked features or attributes derived from them. We design our masking techniques toward capturing different aspects of clinical record data. As clinical records often contain longitudinal data, several of our proposed methods are focused on this scenario, however, we also evaluate a version suitable for static data. For all methods, masking is performed by replacing certain feature values with a fixed value. Therefore, let $F \in \mathbb{R}^{\square x \tau}$ denote a feature matrix and $M \in \mathbb{R}^{\square x \tau}$ denote the corresponding mask, where $\square$ corresponds to the feature dimension of the current feature matrix and $\tau$ to the length of the time-series. For static features $\tau = 1$. The dimensions depend on the considered input features, which can be any of $\mathbf{d},\mathbf{c},\mathbf{t_{d}}$ and $\mathbf{t_{c}}$. $M$ is defined as follows, where $f_{i,j}$ is the feature value at position $(i,j)$:
\begin{equation}
    M_{i,j} = \begin{Bmatrix}
    0\; f_{i,j} \text{ is masked}\\ 
    1 \text{ else}
    \end{Bmatrix}
\end{equation}
For all static features, the masked features $F'$ are then computed as an element-wise product of the mask with the feature matrix (Eq \ref{mask_static}). This leads to masked values being set to zero.
\begin{equation}\label{mask_static}
    F'= F \circ M
\end{equation}
For discrete features, a 'mask token' $m \in \mathbb{N}$ is used to replace masked elements (Eq. \ref{mask_cont}). The mask token $m$ has an arbitrarily chosen, but fixed value. The mask token does not influence the performance, as the discrete masked input is processed by an embedding layer, which functions as a map from feature values to embedding representations.
\begin{equation}\label{mask_cont}
    F'= F \circ M + m * (1-M)
\end{equation}
For time-series features, we further add a binary column per feature to the input vector, that encodes which hours in the time-series are masked. We optimize the model using (binary) cross-entropy loss (\cite{crossentropy}) for the prediction of discrete features and mean squared error loss (\cite{mse}) for predicting continuous features.\\

To mask the feature values we propose different masking strategies for both static as well as time-series input data:

\textbf{Feature Masking}
Given records with an arbitrary number of features for all patients, we randomly mask a fixed percentage of the features, called 'masking ratio', for every record $\mathbf{r_i}$ in the training set. The model is optimized to predict these masked values. The masking ratio is chosen experimentally. This pre-training technique aims to directly teach the model how the patient's features relate to one another, by optimizing it to predict some masked features while having access to the remaining features of the patient. To this end, the masking ratio should be smaller than 100\%, such that only a subset of the features of each patient is masked.

\textit{Static Feature Masking (SFM)}
For static patient data, every patient feature encompasses only one value. Several of these features are randomly selected to be masked and predicted.

\textit{Time-Series Feature Masking (TFM)}
When dealing with time-series features, a completely random selection of feature values to mask would distribute the masked values over the features as well as the time steps. As time-series features in clinical data often encompass repeated or only slowly changing measurements like vital signs or treatment features, feature values of neighboring time points tend to be similar. Knowing the previous and future value of a feature can make predicting random features too easy. An overly simple task has little value for pre-training, as the model does not have to develop a real understanding of the data to solve it. Thus for Time-Series Feature Masking, we mask all time points of an experimentally chosen percentage of each node's features. In this way, the model can neither see previous nor future values to infer the masked feature, but rather has to rely on other features and data from other patients.

\textbf{Block-wise Masking (BM)}
Instead of masking all time points of the given feature time-series, we randomly mask a block of $H$ hours. Again the masking ratio, which determines how many features are masked, is experimentally chosen. Here, the model can access previous and future values of the same feature to make a prediction, but we choose a block size $H > 1$ to avoid knowing all neighboring feature values. The goal of this pre-training method is to teach the model an understanding of the temporal context within a patient's stay.

\textbf{Treatment Prediction (TP)}
Which treatments a patient receives is directly correlated to his condition and its progression. Doctors use the measurements taken from a patient, such as his vital signs and results of disease-specific tests, to decide on the next action in the treatment of the patient. In the health record of a patient the treatments which he receives, e.g. medications that are given, are often recorded. With the Treatment Prediction pre-training, the model is optimized to predict the treatments a patient received, which could help to learn to understand the disease of a patient, given the results of performed measurements. In this task, the model needs to predict a binary label per treatment, indicating if a patient received a certain treatment or not. In case the treatments are given as time-series, we reduce them to binary indicators to generate the labels. In the input data, we mask all treatment features.

\textbf{Patient Masking (PM)}
All previously described pre-training tasks focus on predicting a subset of features of one patient or attributes derived from them. To solve this task it can be sufficient to concentrate on the unmasked feature values of the same patient. With our last pre-training method, we aim to encourage the model to take the other nodes in the graph into account. Toward this goal, we mask a random subset of patients in the graph. For these randomly selected patients, we mask all measurement and treatment features and then again optimize the model to predict these masked features. The only information that remains unmasked is the patient's demographics. Again the percentage of masked patients is decided via experiments.

\textbf{Unsupervised Multi-Task Pre-training}
All of the proposed pre-training strategies have different prediction targets and aim to teach the model distinct knowledge about the graph and the patient features. Thus, combining them during pre-training can lead to a further enhanced and comprehensive understanding of the data. Moreover, training on all tasks simultaneously might have a regularizing effect on the training and can lead to less overfitting. To combine the pre-training tasks, for training we sample one task per batch, mask the samples in the batch correspondingly, make a prediction and then update the model with the corresponding loss. For the next batch, another pre-training task will be sampled. For this combined pre-training task, we need to adapt our decoder to be able to deal with multiple tasks at once. We add a decoder layer for every pre-training task to the model. Depending on which task was sampled the corresponding layer is used and updated. To select the best model we collect all validation predictions of one epoch and average the performance metrics of the different tasks.

All pre-training tasks are sampled with equal probability. Our experiments showed, that even with this straightforward task combination multi-task pre-training is superior on all datasets and tasks. Optimizing the sampling ratio per dataset could potentially improve the effect further, however it would increase the effort needed to apply our method to a new dataset.

\subsubsection{Fine-tuning Step}
After pre-training the model with one of the tasks proposed in the previous section, the pre-trained model is then fine-tuned for a downstream task $T$. 

For \textbf{Self-supervised Learning} the encoder of the model for fine-tuning is initialized using the weights learned during pre-training, while the decoder is initialized randomly. 

For \textbf{Transfer Learning} the initial encoder layers of every feature type (see Figure \ref{arch_overview}) can not be initialized with the pre-trained weights, as the feature dimensions do not match. For example, if the pre-training dataset has N and the fine-tuning dataset M continuous static input features, the linear layer extracting these features needs to have an input dimension of N or M respectively.
Therefore these layers are also initialized randomly, together with the decoder. The rest of the encoder, including the transformer and graph transformer layers, are pre-trained as for self-supervised learning.

In the following, we present our experiments, where we analyze the performance of our model and the effect of the proposed pre-training methods. To this end, we consider the performance improvement in different downstream tasks $T$ after pre-training. We show results for both the self-supervised as well as the transfer learning setting.

\subsubsection{Handling of Missing Data}
When working with EHR data, missing values are very common. They can be caused by lack of collection or documentation, false reporting, or irrelevance of certain values for a given patient (\cite{wells2013strategies}). For time-varying features, which are recorded multiple times over a patient’s stay, another reason for the missingness is, that these features are not measured regularly each hour, but at a different amount of times, reaching from every two hours to once a day. In a data pre-processing step, we impute these missing values, to create a homogeneous input for our network, where every patient has the same amount of features and measurements per feature. In detail, we carry the first and last known value of a time-series forward and backward respectively, while all values in between are linearly interpolated. For features without any value for a patient, we use the mean of this feature from the training set for all time steps. With this technique, the interpolated values provide a meaningful estimation of the missing values. During the pre-training stage, we compute the reconstruction loss solely based on measured values, to avoid optimizing the model to predict possibly incorrect interpolated values. Overall, this enables our method to deal with missing values during the pre-training as well as the fine-tuning stage.

%% file: chapters/experiments_results.tex
\section{Experiments} \label{experiments}
To evaluate our method we use three medical data sets and evaluate with five different downstream tasks. In the following section, we first describe the used datasets and then present and discuss our results and ablation studies.

\subsection{Datasets description:}
We use the two public data sets MIMIC-III (\cite{mimic}) and TADPOLE (\cite{tadpole}), both medical datasets containing multi-modal patient data, for evaluating our method in a self-supervised setting. Additionally, we use the Sepsis Prediction dataset from the PhysioNet/Computing in Cardiology Challenge 2019 \cite{reyna2019early_sepsis_ds, goldberger2000physiobank} to test our model in a transfer learning setup. The datasets are of different sizes and contain different types of input features, allowing for a comprehensive evaluation of our method.

\subsubsection{MIMIC-III} As our first dataset we use MIMIC-III (\cite{mimic}), which is a large dataset, including the EHRs for ICU stays of over 50,000 patients. We use the pre-processed dataset provided by McDermott et al. (\cite{mcdermott2021EHRbenchmark}), which contains the data of ICU stays of 21.88K patients. This data only includes adult patients (at least 15 years old), who stayed one full day (24 hours) or longer in the ICU. Each record contains three types of features:
\begin{itemize}
    \item Demographics: age (static, continuous), gender, admission type, first care unit (static, discrete)
    \item Measurements: 56 different measurements, containing recordings of bedside monitoring and lab test results in hourly granularity (time-series, continuous). Missing values are frequent, as not every measurement is taken every hour. We impute these missing values with linear interpolation in a pre-processing step.
    \item Treatments: binary features per hour, denoting for 16 different treatments, if they were applied in each hour (time-series, static)
\end{itemize}
For every patient, we only use data from the first 24 hours of the stay, as our targeted downstream tasks are defined on this input window. To make our results comparable to previous work, we keep the downstream task definitions and the restriction to the input window of 24 hours. All continuous features in this dataset are normalized to the normal distribution $N(0,1)$. We target three different downstream classification tasks: Length-of-Stay (LOS) and Final Acuity prediction (ACU) as defined in (\cite{mcdermott2021EHRbenchmark}), as well as an adapted acuity prediction task defined in this work. We propose this adapted task because the original acuity prediction task contains a lot of very hard-to-distinguish classes (e.g. Federal Hospital vs Short Term Care Hospital) as well as classes occurring very rarely (down to one sample in the whole dataset). Our adapted task (ACU-4) summarizes the 18 classes of the original task into four meaningful groups. The three downstream tasks are defined as follows:
\begin{description}
\item[Length-of-Stay Prediction (LOS)] For this task, the goal is to predict if a patient will stay longer than three days or not given the data from the first 24 hours of his stay. This can be considered a 2-class classification problem.
\item[Final Acuity Prediction (ACU)] Here the task is to predict a combination of multiple outcomes including the patient's mortality, the discharge location, and the place of death. All together form a multi-class prediction task with 18 classes given below:\\
\textit{Discharge Locations:} Long Term Care Hospital, Rehab/Distinct Part of Hospital, Transfer to Cancer or Children  Hospital,  Home  Health  Care,  Short  Term  Hospital, Intermediate Care Facility (Icf), Transfer to Federal Hospital, Transfer to PsychHospital, Home, Left Against Medical Advice, Home With Home Iv Provider, Other Facility, Hospice-Medical Facility, Skilled Nursing Facility (Snf), Hospice-Home, Snf-Medicaid Only Certified\\
\textit{Death Locations:} In-ICU, In-Hospital
\item[Final Acuity Prediction Adapted (ACU-4)] Again the task is to predict whether the patient will be discharged, and if yes to where, or die. However, we reduced the problem to the following four classes: discharge to home, discharge to home but with additional health care, discharge to care facility, and death.
\end{description}
The total 76 features of MIMIC-III are used as input features, and a subset of them is further used for graph construction. Table \ref{stats_mimic} provides the main statistics about the MIMIC-III dataset and Table \ref{stats_mimic_tasks} about the different tasks.
\begin{table}[htb!]
    \caption{MIMIC-III dataset statistics.}
  \centering
    \begin{tabular}{cccc}
        \toprule
        No. of Samples & No. of Features & \parbox{2.5cm}{\centering No. of Features for Graph}\\
        \midrule
        21.88K & 76 & 56\\
        \bottomrule
    \end{tabular}
    \label{stats_mimic}
\end{table}
\begin{table}[htb!]
    \caption{MIMIC-III task statistics.}
  \centering
    \begin{tabular}{ccccc}
        \toprule
        Task & No. of Classes & Occ. Majority Class\\
        \midrule
        LOS & 2 & 51.81\%\\
        ACU & 18 & 25.24\%\\
        ACU-4 & 4 & 38.37\%\\
        \bottomrule
    \end{tabular}
    \label{stats_mimic_tasks}
\end{table}

\textbf{Graph Construction:}
As MIMIC-III contains a large number of patients, it is computationally infeasible to construct a graph containing all patients and process it with our model. Thus, we randomly split the samples into groups and form sub-graphs with 500 patients each, which fit into memory. Table \ref{graph_size_ablation} shows the performance of our model in the LOS task for different graph sizes. While the performance with 250 and 500 nodes is very similar, using a graph size of 750 nodes leads to a performance drop. This indicates that the restriction to a limited graph size does not harm performance, rather a too-large graph can even have a negative impact. We hypothesize that a too-large graph size can make it harder to learn a meaningful attention distribution as the attention will be distributed over more nodes. At the same time, the gain of relevant new information is limited, as already a large number of similar patients are included in the graph. Every sub-graph contains train, validation, and test patients. By splitting randomly we avoid making assumptions about which patients the model should see at once. Instead, we provide a diverse set of patients in all graphs and allow our model to learn how to attend to these patients via the attention mechanism in the graph transformer.
\begin{table}
\caption{Validation performance of our model with different sub-graph sizes for MIMIC-III. For computational reasons, we performed this test only on one fold.}
\centering
\begin{tabular}{cccc} 
\toprule
& 250 nodes & 500 nodes & 750 nodes \\  
\midrule
AUC & 77.83 & 77.38 & 73.19\\
ACC & 71.10 & 70.96 & 67.85\\
\bottomrule
\end{tabular}
\label{graph_size_ablation}
\end{table}For every subset, we compute the similarity between each pair of patients. As MIMIC-III contains a large number of heterogeneous features, we chose the features for graph construction experimentally. Table \ref{graph_construction_results_mimic} shows the performance in LOS prediction on the first fold of MIMIC-III when using either all or only one specific type of feature for graph construction. Using a graph based on the measurement features proved to be superior, thus we use a similarity computed over the measurement features to construct our population graph. To be able to handle a different amount of measurements in each time-series, we do not directly compare the hourly features, but instead construct feature descriptors $f_d = (mean(f), std(f), min(f), max(f))$ per patient and feature for each of the 56 measurement features. We then compute the average similarity over all feature descriptors $f_d$ between two patients $r_i$ and $r_j$: 
\begin{equation}
Sim(r_i, r_j) = \frac{\sum_{f \in f_d}||f_{r_i} - f_{r_j}||}{|f_d|}
\end{equation}
Given the similarities between all patient pairs for a sub-graph, we construct a k-NN graph with k=5. This leads to on average 18 ± 5.1 disconnected components in the graph.
\begin{table}
\caption{Performance of our model with different graph constructions for MIMIC-III. For computational reasons we performed this test only on one fold.}
\centering
\begin{tabular}{cccccc} 
\toprule
& Age & Treatments & All & \parbox{1.2cm}{\centering Demo-graphics}  & \parbox{1.2cm}{\centering Measure-ments} \\  
\midrule
AUC & 74.79 & 75.20 & 75.20 & 75.21 & \textbf{77.40} \\
ACC & 69.40 & 69.12 & 69.59 & 69.82 & \textbf{71.19} \\ 
\bottomrule
\end{tabular}
\label{graph_construction_results_mimic}
\end{table}

\textbf{Pre-Training Configuration:}
On MIMIC-III, we mask measurement and treatment data from the first 24 hours of each patient's stay. As described before, we compute the loss only over measured features, which are inherently included in the data, and exclude the missing values, which were interpolated in the pre-processing step. We compare all of our proposed pre-training methods, which are designed to handle time-series and treatment data. The masking configurations are given as follows:\\
Time-Series Feature Masking (TFM): masking of 30\% of the features;\\
Block-wise Masking (BM): masking of six-hour blocks in 100\% of the features;\\
Treatment Prediction (TP): all treatments are masked;\\
Patient Masking (PM): masking of 10\% of the patients;\\
Multi-Task (MT): We combine all four tasks (TFM, BM, TP, and PM). For all tasks, we use the same configuration as when trained alone.
For all pre-training tasks, we determined the masking ratios experimentally pre-training with several masking ratios and fine-tuning on LOS prediction. For the further tasks on MIMIC-III, we keep the same masking ratios to be able to use the same pre-trained model. The results of this experiment are shown in Table \ref{mr_mimic}.
\begin{table}[htb!]
  \caption{Experiments to find the optimal masking ratio for all pre-training configurations on MIMIC-III. We fine-tune the pre-trained model with a label ratio of 10\% on the first fold of the MIMIC-III dataset and report validation AUC results in the LOS task. For computational reasons we performed this experiment only over one fold, but as MIMIC-III is a large dataset the findings are general enough as guidance for parameter selection.}
  \centering
    \begin{tabular}{cccccc}
        \toprule
        ratio & 0.15 & 0.3 & 0.5 & 0.75 & 1.0 \\
        \midrule
        BM & 71.48 & 72.37 & 74.30 & 73.92 & \textbf{74.46} \\
        \midrule 
        TFM & 74.29 & \textbf{75.33} & 74.86 & - & -\\
        \midrule
        \midrule 
        ratio & 0.05 & 0.1 & 0.2 & 0.3 & 0.4 \\
        \midrule
        PM & 74.84 & \textbf{76.83} & 75.65 & 75.74 & 75.82\\
        \bottomrule
    \end{tabular}
    \label{mr_mimic}
\end{table}

\subsubsection{TADPOLE} Additionally to MIMIC-III, which is a classical EHR dataset, we evaluate our method on the TADPOLE dataset (\cite{tadpole}), extending our method to multi-modal clinical data. Moreover, this allows us to test our method on a less complex dataset including only static data. This data was obtained from the Alzheimer’s Disease Neuroimaging Initiative (ADNI) database (adni.loni.usc.edu). The ADNI was launched in 2003 as a public-private partnership, led by Principal Investigator Michael W. Weiner, MD. The primary goal of ADNI has been to test whether serial magnetic resonance imaging (MRI), positron emission tomography (PET), other biological markers, and clinical and neuropsychological assessment can be combined to measure the progression of mild cognitive impairment (MCI) and early Alzheimer’s disease (AD). TADPOLE contains a subset of ADNI comprising data from visits of 564 patients.  For each patient, we use a restricted set of 12 features, which were claimed to be informative by the TADPOLE challenge. They include demographics (age, gender, and occurrence of apoe4 gene), results of four cognitive tests, and five imaging features, which are extracted from MR and PET imaging. While the demographics and cognitive test results are discrete, the imaging features are continuous. We normalize the continuous features between zero and one. As a downstream task, we perform diagnosis prediction, classifying each patient into one of the following groups: Cognitive Normal (CN), Mild Cognitive Impairment (MCI), or Alzheimer's Disease (AD). We only use data from patients' first visits to avoid leakage of information. Table \ref{stats_tadpole} provides some basic statistics about the TADPOLE dataset and Table \ref{occ_tadpole} the class occurrences for disease prediction.

\begin{table}[htb!]
    \caption{TADPOLE dataset statistics.}
  \centering
    \begin{tabular}{ccccc}
        \toprule
        Task & \parbox{1.2cm}{\centering No. of Samples} & \parbox{1.1cm}{\centering No. of Feature} & \parbox{1.4cm}{\centering No. of Features for Graph} & \parbox{1.1cm}{\centering No. of Classes}\\
        \midrule
        Disease Pred. & 564 & 12 & 12 & 3\\
        \bottomrule
    \end{tabular}
    \label{stats_tadpole}
\end{table}
\begin{table}[htb!]
    \caption{TADPOLE class occurrences.}
  \centering
    \begin{tabular}{cccc}
        \toprule
        CN & MCI & AD\\
        \midrule
        28.55\% & 56.56\% & 14.89\%\\
        \bottomrule
    \end{tabular}
    \label{occ_tadpole}
\end{table}

\textbf{Graph Construction:}
To construct a patient population graph on TADPOLE, we compute a feature similarity between all patients. As we work with a reduced feature set of only 12 features, we decided to use all features to compute the feature similarity. Dependent on the feature type this similarity is computed differently. For the demographics, age, gender, and apoe4, we use absolute feature matching, checking if patients have the same gender/ apoe4 feature or are of similar age:
\begin{equation}
S_{dem}(r_i, r_j) = \sum \begin{cases}   
1 \textnormal{ if } f_i = f_j  \textnormal{ else } 0 \\  
1 \textnormal{ if } \left|age_{i} - age_{j} \right| \le 2 \textnormal{ else } 0
\end{cases}\div 3 \textnormal{,}
\end{equation}
where \textit{f}=(apoe4, gender).

For the cognitive test results $\mathbf{d_i}$, which are discrete but ordinal features, and the continuous imaging features $\mathbf{c_i}$, we calculate the respective normalized L2 distances:
\begin{equation}
S_{cog}(r_i, r_j) = \frac{\sum_{f \in \mathbf{d_i}}||f_{r_i} - f_{r_j}||}{max(\mathbf{d_i})}
\end{equation}
\begin{equation}
S_{img}(r_i, r_j) = sig({\sum_{f \in \mathbf{c_{i}}}||f_{r_i} - \\f_{r_j}||}).
\end{equation}

Given all these similarities, we construct a k-NN graph with k=5, dependent on the mean similarity ($S$), which is computed as the mean of $S_{dem}$, $S_{cog}$ and $S_{img}$. This leads to on average 6.1 ± 1.1 separated components in the graph.

\textbf{Pre-Training Configuration:} As TADPOLE is a less complex dataset and does not include time-series data, we apply only Static Feature Masking as a pre-training task. For this, we apply masking to the APOE4 gene feature, the cognitive test results, and the imaging features with a masking ratio of 30\%. We set the masking ratio experimentally, as shown in Table \ref{mr_tadpole}. Although Static Feature Masking is a simple masking technique, this experiment allows us to evaluate the benefit of our overall framework on static, multi-modal, clinical data, including the modeling as a patient population graph and the combination of masking-based pre-training with a transformer-based architecture.

\begin{table}[htb!]
  \caption{Experiments to find the optimal masking ratio for pre-training on TADPOLE. We fine-tune with a label ratio of 1\% and evaluate with 10-fold cross-validation.}
  \small
  \centering
    \begin{tabular}{ccccccc}
        \toprule
        ratio & 0.1 & 0.2 & 0.3 & 0.4 \\
        \midrule
        AUC & 91.45$\pm$3.11 & 93.07$\pm$2.29 & \textbf{93.49$\pm$2.07} & 93.37$\pm$1.95 \\
        \bottomrule
    \end{tabular}
    \label{mr_tadpole}
\end{table}

\subsubsection{Sepsis Prediction Dataset}
We work with the Sepsis Prediction dataset from the PhysioNet/Computing in Cardiology Challenge 2019 (\cite{reyna2019early_sepsis_ds, goldberger2000physiobank}). It includes longitudinal EHR data, including vital signs and laboratory values, from patient stays in two different hospitals, as well as the patient's demographics.

The publicly available part of the Sepsis Prediction dataset includes 40,336 ICU patients, equally distributed between two hospitals. The features of every patient include eight vital signs, 26 laboratory values, and six demographics. As for MIMIC-III, the measurement features (vital signs and laboratory values) are time-series features with hourly granularity and a high ratio of missing values. Table \ref{stats_sepsis} provides some basic statistics about the Sepsis Prediction dataset.

For every patient, an hour-wise label indicates if the patient has sepsis. In this work, we target a patient-level binary classification instead of an hourly prediction and predict if a patient will develop sepsis or not. We crop the data of each patient such that for septic patients we only see a time window of 24 hours ending six hours before the onset of the sepsis. For non-septic patients, we select a random window of 24 hours. Thus we still task the model to identify sepsis before its onset, but we concentrate on the time window close to onset. Further we apply the same data pre-processing as in the MIMIC-III dataset, including the normalization of continuous features to the normal distribution and interpolation of missing values.

The graphs for the Sepsis Prediction dataset are created analogously to MIMIC-III. Here this results in 17.7 ± 4.1 components per graph.

\begin{table}[htb!]
  \centering
    \begin{tabular}{ccccc}
        \toprule
        Task & \parbox{1.2cm}{\centering No. of Samples} & \parbox{1.2cm}{\centering No. of Features} & \parbox{1.2cm}{\centering No. of Classes} & \parbox{1.2cm}{Occ. Sepsis}\\
        \midrule
        Sepsis Pred. & 40336 & 40 & 2 & 7.27\%\\
        \bottomrule
    \end{tabular}
    \caption{Sepsis Prediction dataset statistics.}
    \label{stats_sepsis}
\end{table}

\textbf{Comparison to MIMIC-III}
Like MIMIC-III, the Sepsis Prediction dataset contains longitudinal measurement features and patient demographics. For the demographics, both datasets contain age and gender, the other demographics are different. From the measurements in the Sepsis Prediction dataset, 24 features are also included in MIMIC-III, while 10 features are new. Further, MIMIC-III contains additional measurement features that are not included in the Sepsis Prediction dataset. Moreover, the Sepsis Prediction dataset does not contain any treatment features.

\subsection{Implementation Details}
All experiments are implemented in PyTorch and performed on a TITAN Xp GPU with 12GB VRAM. For cross-validation, pre-training is performed separately on the training data of every fold to avoid leaking information from the validation or test sets during pre-training. All experiments are optimized using the Adam optimizer (\cite{kingma2014adam}). We manually tuned hyperparameters per dataset separately for each pre-training task and for from scratch training as well as for fine-tuning of each downstream task. \\
\textbf{MIMIC-III:} The backbone for the MIMIC-III model consists of $\mathbf{L} = 8$ Graphormer layers. The best model is selected given the performance on the validation set, and all models are trained until convergence of the validation performance. For a fair comparison with the state of the art, all results are averaged over six folds as provided by \cite{mcdermott2021EHRbenchmark}, each with an 80-10-10 split into train, validation, and test data, and computed over the respective test sets. The learning rates for the different pre-training and fine-tuning tasks on MIMIC-III are given in Tables \ref{mimic_lrs_pt} and \ref{mimic_lrs_task}. In Table \ref{mimic_lrs_pt} we denote the use of a polynomial decaying learning rate as $start\_lr - end\_lr$.
\begin{table}[htb!]
\centering
\begin{tabular}{cccccc}
\toprule
       & TFM   & BM   & TP   & PM & MT  \\ \midrule
lr     & 1e-3 - 1e-4 & 1e-3 - 1e-4 & 5e-4 & 5e-4 & 5e-4 \\ \bottomrule
\end{tabular}
\caption{Learning rates for pre-training on MIMIC-III.}
\label{mimic_lrs_pt}
\end{table}

\begin{table}[htb!]
\centering
\begin{tabular}{c|cc|cc|cc}
\toprule
Task                        & \multicolumn{2}{c|}{LOS}          & \multicolumn{2}{|c|}{ACU} & \multicolumn{2}{|c}{ACU-4} \\ \midrule
       & SC   & FT   & SC          & FT         & SC          & FT          \\ \midrule
lr     & 1e-4 & 1e-5 & 1e-4        & 1e-5       & 1e-4        & 1e-5/5e-5*        \\\bottomrule
\end{tabular}
\caption{Learning rates for from scratch training (SC) and fine-tuning (FT) on the MIMIC-III dataset. *5e-5 when pre-training with Patient Masking, else 1e-5}
\label{mimic_lrs_task}
\end{table}
\textbf{TADPOLE:} The backbone for the TADPOLE model consists of $\mathbf{L} = 4$ Graphormer layers. For pre-training, we train the model with a learning rate of 1e-5 for 6000 epochs. For from scratch training on the downstream task, we use a polynomial decaying learning rate, which is reduced from 1e-5 to 5e-6 for 1200 epochs. To fine-tune pre-trained models, we use a learning rate of 5e-6 and perform training again for 1200 epochs. When training with a label ratio of 1\% we reduce the epochs to 200, as for this small amount of data, the pre-trained models reach optimal performance much faster than other models. All results are computed using 10-fold cross-validation.\\
\textbf{Sepsis Prediction dataset:}
We use the same model as for the MIMIC-III dataset with $\mathbf{L} = 8$ Graphormer layers. The encoder is adapted to fit the Sepsis Prediction dataset's features. We perform 5-fold cross-validation, where every rotation is split into train, validation, and test sets with an 80-10-10 split, and select the best model based on the validation set performance per split. We use a learning rate of 1e-4 for both from scratch training and transfer learning.

\subsection{Results and Discussion}
In our experiments, we aim to evaluate the proposed model on different downstream tasks, compare the performance of our model to related work and evaluate the performance improvement that can be reached with our pre-training methods. Further, we compare different pre-training methods to one another and provide various ablation studies, investigating the importance of the different modules of our proposed model architecture.
\subsubsection{Comparative methods}
To show the positive impact of our (Graph)-Transformer-based architecture, we compare our model to related work without any pre-training. Therefore we train our model from scratch on the full training data of MIMIC-III and TADPOLE for different downstream tasks and compare the results to previous work. The results are shown in Table \ref{full_data_results}.

For MIMIC-III, both targeted downstream tasks, LOS and ACU, were defined in the EHR pre-training benchmark by \cite{mcdermott2021EHRbenchmark}. Similar tasks were targeted by previous work, but only a few works target the same tasks on the MIMIC-III dataset. As we use the exact task definitions by McDermott et al. and use the pre-processed dataset they provide, comparing with their work is most meaningful. Further, we compare to MIMIC-Extract (\cite{wang2020mimicextract}) for LOS. They compared different models for 3-day length-of-stay prediction. We report the performance of the best-performing model, which is a random forest classifier. ACU prediction was so far only targeted by McDermott et al. and our newly defined ACU-4 task was not targeted before, which is why we can not provide results of previous work. We outperform the random forest model by Wang et al. in LOS as well as the GRU-based model by McDermott et al. in LOS and ACU prediction, showing the benefit of modeling MIMIC-III as a population graph as well as our model architecture (Table \ref{full_data_results}).

For the TADPOLE dataset, the most important previous works and state-of-the-art models in diagnosis prediction on TADPOLE are DGM (\cite{kazi2022differentiable}) and a latent graph learning framework proposed by \cite{cosmo2020latent}. Similar to our work, they model the data as a population graph, but instead of pre-defining this graph, the proposed models learn a population graph in an end-to-end manner for a given downstream task. This allows the model to decide which patients should be connected depending on the current task. Even though our population graph is fixed, our model still has the freedom to weigh the importance of other patients by learning global attention towards all nodes in the graph. Besides, one recent arxiv paper (\cite{kazi2021ia}) could further improve disease classification performance on TADPOLE by learning the importance of the given input features. However, it is out of context for this work. We outperform the work by \cite{kazi2022differentiable} and reach similar accuracy as \cite{cosmo2020latent} while outperforming in AUC, which is the more important metric as TADPOLE is an imbalanced dataset (Table \ref{full_data_results}).

Overall, we show significant performance gains on the MIMIC-III dataset and achieve comparable results on TADPOLE. The good results on both MIMIC-III and TADPOLE show that the proposed data modeling and model architecture are a good fit for the task at hand.
\begin{table}[htb!]
     \caption{Accuracy and AUC of the proposed method compared with DGM on TADPOLE and the model proposed by \cite{mcdermott2021EHRbenchmark} on MIMIC-III.}
  \centering
    \begin{tabular}{ccc}
        \toprule
        Model & ACC & AUC \\
        \midrule
        \multicolumn{3}{c}{MIMIC-III: LOS}\\
        \midrule 
        \cite{wang2020mimicextract} - RF & 68.3 & 73.3\\
        \cite{mcdermott2021EHRbenchmark} & not reported & 71.00 $\pm$ 1.00\\
        Proposed & \textbf{70.29 $\pm$ 1.10} & \textbf{76.17 $\pm$ 1.02}\\
        \midrule 
        \multicolumn{3}{c}{MIMIC-III: ACU}\\
        \midrule 
        \cite{mcdermott2021EHRbenchmark} & not reported & 78.00 $\pm$ 2.00\\
        Proposed & 43.67$\pm$0.83 & \textbf{79.42$\pm$2.72}\\
        \midrule 
        \multicolumn{3}{c}{TADPOLE} \\
        \midrule 
        \cite{cosmo2020latent} & \textbf{92.91 $\pm$ 02.50} & 94.49 $\pm$ 03.70\\
        \cite{kazi2022differentiable} & 91.05 $\pm$ 5.93 & 96.86 $\pm$ 1.81\\
        Proposed & 92.59 $\pm$ 3.64 & \textbf{96.96 $\pm$ 2.32}\\
        \bottomrule
    \end{tabular}
    \label{full_data_results}
\end{table}

\subsubsection{Effect of pre-training}
The focus of our work lies in investigating the benefit of pre-training for patient-level prediction tasks, especially for scenarios with limited labeled data. To evaluate our proposed pre-training techniques, we compare the performance of our model trained from scratch to our pre-trained and subsequently fine-tuned model. In general, pre-training is especially suitable to be applied in scenarios where much more data is available for pre-training than for fine-tuning. To simulate this case, we artificially create scenarios with limited labeled data, by fine-tuning the model only with a subset of the given labels. We vary the label ratio, meaning the number of labels used for fine-tuning, between 1\%, 5\%, 10\%, 50\%, and 100\%. On the other hand, we always use 100\% of the given data for unsupervised pre-training. The results emphasize the benefits of our unsupervised pre-training with limited labels.

\textbf{MIMIC-III} is a complex dataset, including time-series features and multiple challenging downstream tasks. This allows us to evaluate and compare all of our proposed pre-training strategies on multiple prediction tasks.

\begin{figure*}[htb!]
\centering
\includegraphics[width=\linewidth]{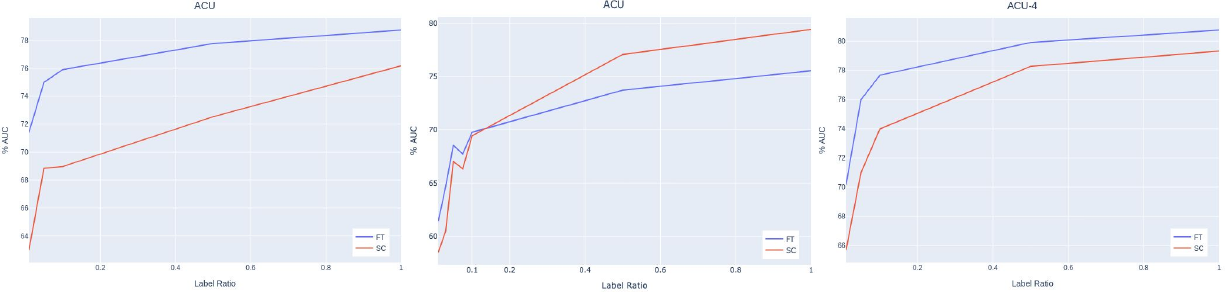}
\caption{Visualization of the AUC development comparing from scratch and multi-task pre-training on all tasks on MIMIC-III. The improvement decreases with higher label ratios. For the ACU task pre-training only improves performance for small label ratios up to 10\%.}
\label{development_los}
\end{figure*}

\begin{table*}[htb!]
    \caption{Performance of the proposed model in accuracy and AUC trained from scratch (SC) or fine-tuned after pre-training (FT) with the different proposed pre-training tasks (TP, BM, TFM, PM, MT) for different label ratios for the LOS task.}
  \centering
    \begin{tabular}{ccccccc|c} 
        \toprule
        \multicolumn{8}{c}{MIMIC-III: LOS} \\
        \midrule
        Labels & Metric & SC & FT: TP & FT: BM & FT: TFM & FT: PM & FT: MT\\
        \midrule
        1\% & ACC & 59.86 $\pm$ 2.11 & 64.40 $\pm$ 1.17 & 63.22 $\pm$ 2.39 & 65.25 $\pm$ 1.09 & \color{teal}65.72 $\pm$ 1.53&  \textbf{66.39 $\pm$ 1.34}\\
        & AUC & 62.98 $\pm$ 2.55 & 68.23 $\pm$ 1.18 & 68.07 $\pm$ 1.80 & 69.90 $\pm$ 1.26 & \color{teal}70.84 $\pm$ 2.59&  \textbf{71.40 $\pm$ 1.81}\\
        \midrule
        5\% & ACC & 64.79 $\pm$ 1.16 & 66.23 $\pm$ 0.88 & 66.82 $\pm$ 0.89 & \color{teal}68.66 $\pm$ 0.73 & 68.41 $\pm$ 0.92&  \textbf{68.97 $\pm$ 0.46}\\
        & AUC & 68.85 $\pm$ 1.53 & 70.99 $\pm$ 0.03 & 72.27 $\pm$ 1.19 & 73.97 $\pm$ 1.28 & \color{teal}74.31 $\pm$ 1.04&  \textbf{74.99 $\pm$ 1.00}\\
        \midrule
        10\% & ACC & 64.72 $\pm$ 0.45 & 66.92 $\pm$ 1.10 & 67.71 $\pm$ 0.69 & \color{teal}69.42 $\pm$ 1.23 & 69.19 $\pm$ 0.58&  \textbf{69.99 $\pm$ 0.97}\\
        & AUC & 68.97 $\pm$ 0.66 & 71.57 $\pm$ 0.99 & 73.55 $\pm$ 0.60 & \color{teal}75.09 $\pm$ 1.29 & 74.92 $\pm$ 0.87&  \textbf{75.90 $\pm$ 1.35}\\
        \midrule
        50\% & ACC & 67.41 $\pm$ 1.31 & 68.87 $\pm$ 0.89 & 69.98 $\pm$ 0.69 & \color{teal}70.85 $\pm$ 0.92 & 70.71 $\pm$ 1.01&  \textbf{71.46 $\pm$ 0.91}\\
        & AUC & 72.53 $\pm$ 1.08 & 74.21 $\pm$ 1.11 & 76.02 $\pm$ 0.87 & 76.86 $\pm$ 1.47 & \color{teal}77.03 $\pm$ 0.95&  \textbf{77.77 $\pm$ 1.17}\\
        \midrule
        100\% & ACC & 70.29 $\pm$ 1.10 & 69.84 $\pm$ 1.11 & 70.73 $\pm$ 0.70 & \color{teal}71.44 $\pm$ 1.25 & 71.02 $\pm$ 0.73&  \textbf{72.13 $\pm$ 1.46}\\
        & AUC & 76.17 $\pm$ 1.02 & 75.47 $\pm$ 0.86 & 76.20 $\pm$ 0.54 & 77.78 $\pm$ 1.31 & \color{teal}77.99 $\pm$ 0.90&  \textbf{78.73 $\pm$ 1.21}\\
        \bottomrule
    \end{tabular}
    \label{results_table_los}
\end{table*}
\begin{table*}[htb!]
\caption{ Performance of the proposed model in accuracy and AUC trained from scratch (SC) or fine-tuned after pre-training (FT) with the different proposed pre-training tasks (TP, BM, TFM, PM, MT) for different label ratios on the ACU task.}
\centering
\begin{tabular}{ccccccc|c}
\toprule
\multicolumn{8}{c}{MIMIC-III: ACU}    \\ \midrule
Labels  & Metric & SC                                                        & FT:TP & FT:PM  & FT:BM   & FT:TFM  & FT:MT           \\ \midrule
1\%   & ACC    & 28.10$\pm$2.94   & 30.019$\pm$4.30 & 32.25$\pm$1.59 & {\color{teal} 33.31$\pm$2.19} & \textbf{34.30$\pm$2.73}& 33.17$\pm$1.96        \\
      & AUC    & 58.51$\pm$0.80   & 59.39$\pm$2.41 & {\color{teal}60.86$\pm$0.94}& 59.59$\pm$2.70 & 60.10$\pm$2.09 & \textbf{61.43$\pm$1.86} \\ \midrule
5\%   & ACC    & 35.16$\pm$1.88   & 36.76$\pm$1.04 & 39.06$\pm$1.20  & 39.77$\pm$1.05 & {\color{teal}40.57$\pm$0.97} & \textbf{40.86$\pm$0.79}      \\
      & AUC    & 67.06$\pm$2.84   & 63.22$\pm$2.23 & 66.56$\pm$1.33  & 66.93$\pm$2.25  & {\color{teal}67.55$\pm$1.58} & \textbf{68.58$\pm$1.27} \\ \midrule
10\%  & ACC    & 36.69$\pm$1.06   & 36.86$\pm$3.49 & 40.71$\pm$0.75  &{\color{teal} 41.93$\pm$0.95}  & \textbf{42.47$\pm$1.01} & 41.91$\pm$1.15                   \\
      & AUC    & {\color{teal}69.44$\pm$2.55}   & 64.28$\pm$3.08 & 66.43$\pm$1.86  & 68.38$\pm$2.35 &  69.13$\pm$0.91 & \textbf{69.77$\pm$1.98} \\ \midrule
50\% & ACC    & 42.68$\pm$1.57   & 43.52$\pm$1.20 & 42.98$\pm$1.29 &  44.81$\pm$0.23 & {\color{teal}44.87$\pm$0.98}   & \textbf{45.39$\pm$0.71}                  \\
      & AUC    & \textbf{77.07$\pm$2.22}   & 72.30$\pm$2.09 & 68.88$\pm$3.15   & 72.35$\pm$3.95 & {\color{teal} 73.78$\pm$1.83} & 73.72$\pm$2.39 \\ \midrule
100\% & ACC    & 43.67$\pm$0.83   & 44.71$\pm$1.06 & 43.78$\pm$1.34  & 45.47$\pm$0.58  & \textbf{46.27$\pm$0.76}& {\color{teal}46.26$\pm$0.85}                      \\
      & AUC    & \textbf{79.42$\pm$2.72}  & 73.75$\pm$2.00 & 70.66$\pm$2.60  & {\color{teal} 75.60$\pm$3.48} & 75.42$\pm$1.66 & 75.54$\pm$2.35 \\ \midrule
\end{tabular}
\label{results_table_acu}
\end{table*}
\begin{table*}[htb!]
	\caption{Performance of the proposed model in accuracy and AUC trained from scratch (SC) or fine-tuned after pre-training (FT) with the different proposed pre-training tasks (TP, BM, TFM, PM, MT) for different label ratios for the ACU-4 task. The best and second-best results per label ratio of all single-task pre-training methods and from-scratch training are marked as bold/green.}
	\centering
	\begin{tabular}{ccccccc|c}
		\toprule
		\multicolumn{8}{c}{MIMIC-III: ACU-4} \\
		\midrule
		Labels  & Metric & SC                            &FT: TP & FT: PM        & FT: BM                                          & FT: TFM    &    FT: MT  \\ 
		\midrule
		1\%   & ACC    & 41.03 $\pm$ 2.86   & 43.10 $\pm$ 2.61& 45.16 $\pm$ 2.58 & 44.78 $\pm$ 3.14 & \color{teal}46.23 $\pm$ 2.79 &  \textbf{46.24 $\pm$ 2.31}  \\
		& AUC    & 65.67 $\pm$ 1.95    & 65.70 $\pm$ 1.27& 68.13 $\pm$ 1.76 & 68.40 $\pm$ 3.07  & \textbf{70.21 $\pm$ 2.17} &  \color{teal}70.16 $\pm$ 1.97 \\
		\midrule
		5\%   & ACC    & 46.33 $\pm$ 2.05   & 48.10 $\pm$ 0.71& 50.49 $\pm$ 1.27 & 50.71 $\pm$ 1.16  & \color{teal}51.77 $\pm$ 0.63  & \textbf{52.68 $\pm$ 1.51}\\
		& AUC    & {71.00 $\pm$ 1.17}   & 70.84 $\pm$ 0.95& 73.59 $\pm$ 1.09 & 75.09 $\pm$ 0.65  & \color{teal}75.62 $\pm$ 0.72 &  \textbf{76.01 $\pm$ 1.10} \\
		\midrule
		10\%  & ACC    & 49.93 $\pm$ 1.02    & 50.71 $\pm$ 0.82& 51.63 $\pm$ 0.75 & 52.82 $\pm$ 1.48 & \color{teal}53.77 $\pm$ 1.27 &  \textbf{54.41 $\pm$ 1.73} \\
		& AUC    & 73.99 $\pm$ 0.64 & 73.07 $\pm$ 0.28& 75.17 $\pm$ 1.06 & 76.90 $\pm$ 0.91 & \color{teal}77.09 $\pm$ 0.71&  \textbf{77.67 $\pm$ 0.79} \\ \midrule
		50\%  & ACC    & 54.02 $\pm$ 1.27& 54.42 $\pm$ 1.16& 53.76 $\pm$ 1.40 & 55.93 $\pm$ 1.28 & \color{teal}56.17 $\pm$ 1.18  &  \textbf{56.30 $\pm$ 1.27} \\
		& AUC    & 78.28 $\pm$ 0.56   & 77.69 $\pm$ 0.65& 77.70 $\pm$ 0.98 & 79.63 $\pm$ 0.63  & \color{teal}79.66 $\pm$ 0.60 &  \textbf{79.92 $\pm$ 0.70} \\ 
		\midrule
		100\% & ACC    & 55.41 $\pm$ 0.87   & 54.98 $\pm$ 1.20& 54.75 $\pm$ 1.03 & \color{teal}56.86 $\pm$ 1.26  & 56.84 $\pm$ 0.71 &  \textbf{57.11 $\pm$ 1.19} \\
		& AUC    & 79.33 $\pm$ 0.72 & 78.91 $\pm$ 0.76& 78.72 $\pm$ 0.90 & 80.44 $\pm$ 0.67 & \color{teal}80.51 $\pm$ 0.45  & \textbf{ 80.77 $\pm$ 0.78} \\ 
		\bottomrule
	\end{tabular}
	\label{results_table_acu4}
\end{table*}
For LOS the results are shown in table \ref{results_table_los}. Further, the development of the improvement through pre-training for different label ratios is visually shown in figure \ref{development_los} for all tasks on MIMIC-III. For LOS, both accuracy and AUC improve significantly for all label ratios compared to from scratch training, independent of the used pre-training task. When comparing the different pre-training tasks (TM, BM, TFM, and PM), it can be observed that Feature and Patient Masking consistently outperform Block-wise Masking and Treatment Prediction. Further Treatment Prediction overall shows the least positive effect. Considering the difficulty of the pre-training tasks, the easier tasks tend to be less beneficial than the harder ones. Treatment Prediction only asks the model to predict binary indicators if a certain treatment was applied, while the Feature Masking task includes predicting time-series features describing the treatment application in each hour. So conceptually Treatment Prediction is a sub-task of Feature Masking. Similarly during Block-wise Masking the model has to predict feature values only for a few hours while having access to past and future measurements, which makes the task simpler compared to predicting a fully masked feature as in Feature Masking. These results indicate that a certain level of complexity is beneficial for a valuable pre-training task.

For acuity (ACU) prediction we consider both the original as well as our adapted ACU-4 task. However, during our experiments we observed that many of the 18 classes from the original task are not predicted at all, neither when training from scratch nor when fine-tuning. This supports our claim, that the task is not well defined including missing, very rare, and very similar classes. Therefore, we introduce our newly defined ACU-4 task to get more meaningful results.

For ACU prediction, pre-training improves both metrics for small label ratios (1\% till 10\%) as shown in Table \ref{results_table_acu}. For higher label ratios only accuracy improves. Again Feature Masking consistently outperforms Block-wise Masking and Treatment prediction shows the least performance improvements. When training on the adapted ACU-4 task, a clear benefit of all pre-training methods can be observed (see Table \ref{results_table_acu4}). Similar to the LOS task, Treatment Prediction is the least beneficial and Feature Masking is the most beneficial, supporting our hypothesis that the difficulty of the pre-training tasks influences their effect. However, for both acuity prediction tasks (ACU and ACU-4) Patient Masking shows less benefit than Block-wise and Feature Masking, indicating that the information of other patients is less relevant for acuity prediction than for length-of-stay prediction. Generally, Patient Masking is a complex task, but it requires the information of other patients to be helpful in solving the end task. Therefore the value of Patient Masking is dependent on the correlation between different patients, which can differ depending on the targeted task. This also shows that the selection of the optimal pre-training task depends on the current downstream task.

Multi-Task Pre-training further improves the performance after fine-tuning for all downstream tasks, LOS, ACU, and ACU-4. This indicates that the different tasks convey at least partially complementary knowledge and together lead to an improved understanding of the data. Moreover, as described previously the effectiveness of the single pre-training tasks differs between tasks, showing that the selection of the best pre-training task is dependent on the downstream task. With Multi-Task Pre-training this decision can be circumvented, as it generally outperforms the best of the single methods. Consequently, Multi-Task Pre-training is well suited as an out-of-the-box pre-training method, as it is less dependent on the downstream task.

\begin{figure}[htb!]
\centering
\includegraphics[width=\linewidth]{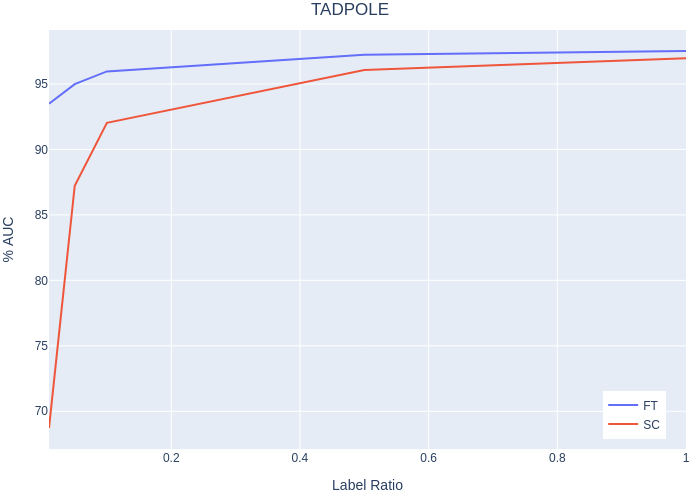}
\caption{Visualization of the AUC development comparing from scratch and pre-training on TADPOLE. Again the improvement decreases with higher label ratios.}
\label{development_tadpole}
\end{figure}

\textbf{TADPOLE}
Additionally, we use the less complex TADPOLE dataset to evaluate our method on multi-modal, but static clinical data. The results of the fine-tuned model, which was pre-training with Static Feature Masking, as well as the model trained from scratch for disease prediction are shown in Table \ref{results_table_tadpole} and the performance development in relation to the label ratio is visualized in figure \ref{development_tadpole}. 
\begin{table}[htb!]
    \caption{Performance of the proposed model in accuracy and AUC trained from scratch (SC) or fine-tuned after pre-training (FT) for different label ratios on the TADPOLE dataset.}
  \centering
    \begin{tabular}{cccc}
        \toprule
         \multicolumn{4}{c}{TADPOLE} \\
        \midrule
        Labels & Metric & SC & FT\\
        \midrule
        1\% & ACC & 59.42 $\pm$ 8.40 & \textbf{78.89 $\pm$ 2.45}\\
        & AUC & 68.72 $\pm$ 12.74 & \textbf{93.49 $\pm$ 2.07}\\
        \midrule
        5\% & ACC & 78.23 $\pm$ 6.83 & \textbf{83.37 $\pm$ 6.29}\\
        & AUC & 87.23 $\pm$ 4.91 & \textbf{94.99 $\pm$ 2.55}\\
        \midrule
        10\% & ACC & 87.00 $\pm$ 4.86 & \textbf{87.71 $\pm$ 4.65}\\
        & AUC & 92.03 $\pm$ 3.39 & \textbf{95.96 $\pm$ 2.51}\\
        \midrule
        50\% & ACC & \textbf{92.41 $\pm$ 3.69} & 91.52 $\pm$ 3.76\\
        & AUC & 96.06 $\pm$ 2.48 & \textbf{97.23 $\pm$ 1.94}\\
        \midrule
        100\% & ACC & \textbf{92.59 $\pm$ 3.64} & 92.24 $\pm$ 3.47\\
        & AUC & 96.96 $\pm$ 2.23 & \textbf{97.52 $\pm$ 1.67}\\
        \bottomrule
    \end{tabular}
    \label{results_table_tadpole}
\end{table}
Even though TADPOLE is a rather small dataset and thus also the amount of data used for pre-training is limited, we show a significant benefit through pre-training. Especially in settings with limited labels (1\%, 5\%, 10\%) pre-training helps to improve performance significantly. While the model trained from scratch starts to overfit very fast when trained with very few samples (e.g. around 50 samples for a label ratio of 1\%), the pre-trained model reaches a quite good performance, by quickly adapting to the provided samples. Moreover, the AUC continues to improve when pre-training the model up to the full dataset size, showing that pre-training helps to deal with the class imbalance in this dataset. This experiment shows the value of our method for static, multimodal clinical data, and the benefit of pre-training when labeled data is very limited.

When analyzing the confusion matrices for models trained from scratch and pre-trained models, we can conclude that the major improvement through pre-training in binary tasks (LOS) lies in reducing false negatives. For multi-class tasks such as on TADPOLE or ACU-4 prediction on MIMIC-III, false negatives are mainly reduced for the minority classes while for the majority class false positives are reduced. Overall, this shows that pre-training helps the model to miss fewer positive cases, especially for minority classes.

\medskip
For the MIMIC-III dataset, \cite{mcdermott2021EHRbenchmark} proposed a benchmark for pre-training on EHR data and evaluated their method among others on Length-of-Stay and Acuity Prediction. Table \ref{PT_comparison} shows a direct comparison of the pre-training methods proposed by them and our best-performing method (Multi-Task Pre-training). They propose one unsupervised and one supervised pre-training method. In their unsupervised method (MI), they mask and predict random time-points of the EHR record, while for their supervised method (MT), they perform multi-task training on nine out of ten supervised patient outcome prediction tasks and fine-tune the model on the task which was left out. When comparing to the unsupervised masked imputation pre-training our method is clearly superior, both in terms of overall AUC as well as the performance gain achieved by pre-training, showing that our masking strategies better adhere to the nature of this longitudinal and heterogeneous data than simply masking random time points. In comparison to the supervised multi-task pre-training, our method achieves similar performance gains with 1\% labels but continues to improve performance in Length-of-Stay prediction also for higher label ratios, while the multi-task pre-training starts to degrade performance. The good results in comparison to this method are especially notable, as we do not use any labels for pre-training, while the supervised pre-training considers labels of multiple prediction tasks. Overall, the results indicate that our modeling of the data as a patient population graph together with our pre-training approaches, which are specifically designed for clinical (time-series) data, are better suited for the given scenario. 

Besides the work of McDermott et al., \cite{gupta2020transfer} also propose a method for pre-training on MIMIC-III for phenotype and mortality prediction, but they neither provide an implementation nor sufficient details to re-implement their method from scratch. The results they reported for their unsupervised transfer learning method show only slight improvements in the AUROC score over an LSTM trained from scratch while leading to a slightly worse AUPRC score. For disease prediction on TADPOLE, no previous work proposed a pre-training approach.
\begin{table}[htb!]
 \caption{Comparison to other pre-training methods on MIMIC-III. *MI and MT are the masked imputation and multi-task pre-training proposed by McDermott et al. (\cite{mcdermott2021EHRbenchmark}). MI is an unsupervised pre-training method, while MT is supervised.}
  \centering
    \begin{tabular}{ccccc}
        \toprule
        \multicolumn{5}{c}{LOS} \\
        \midrule
        Labels & Metric & MI* & MT* & Proposed\\
        \midrule
        1\% & AUC & 62 $\pm$ 3 & 67 $\pm$ 3 & \textbf{71.40 $\pm$ 1.81}\\
        & gain & +4 & \textbf{+9} & +8.42 \\\midrule
        10\% & AUC & 60 $\pm$ 2 & 65 $\pm$ 3 & \textbf{75.90 $\pm$ 1.35}\\
        & gain & -6 & -1 & \textbf{+6.93} \\\midrule
        100\% & AUC & 58 $\pm$ 3 & 64 $\pm$ 4 & \textbf{78.73 $\pm$ 1.21}\\
        & gain & -13 & 64 $\pm$ -7 & \textbf{+2.56} \\
        \midrule
        \multicolumn{5}{c}{ACU} \\
        \midrule
        1\% & AUC & 60 $\pm$ 4 & \textbf{60 $\pm$ 1} & \textbf{60.10 $\pm$ 2.09}\\
        & gain & +1 & +1 &\textbf{+1.59} \\\midrule
        10\% & AUC & 69 $\pm$ 3 & \textbf{70 $\pm$ 4} & 69.13 $\pm$ 0.91\\
        & gain & +0 & \textbf{+1} & -0.31 \\\midrule
        100\% & AUC & 74 $\pm$ 2 & 74 $\pm$ 4 & \textbf{75.42 $\pm$ 1.66}\\
        & gain & \textbf{-4.0} & \textbf{-4.0} & \textbf{-4.0} \\
        \bottomrule
    \end{tabular}
    \label{PT_comparison}
\end{table}

Overall we show, that our unsupervised pre-training pipeline helps remarkably to improve performance on both datasets and for multiple patient outcome prediction tasks over patient population graphs. As expected, the highest improvements can be seen in settings with limited labels, however, even when using all labels a notable improvement remains. Further, it can be observed that the pre-trained models mostly have a lower standard deviation than the models trained from scratch, indicating that pre-training also improves model stability. From comparing our different pre-training tasks, we can conclude that the difficulty of the pre-training tasks correlates with their effectiveness.

\subsubsection{Application to Transfer Learning}
\begin{figure}[htb!]
\centering
\includegraphics[width=\linewidth]{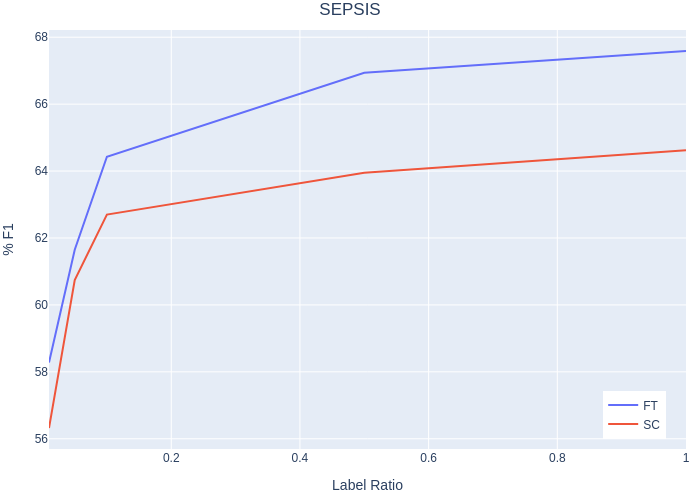}
\caption{Visualization of the F1 development comparing from scratch and multi-task pre-training on the Sepsis prediction dataset. Here the performance improvement remains nearly constant for different label ratios.}
\label{development_sepsis}
\end{figure}

To evaluate our method in a transfer learning setting, we pre-train our model with multi-task pre-training on MIMIC-III and then fine-tune it on the Sepsis Prediction dataset, which provides overlapping but different features per patient. The results of this experiment are shown in Table \ref{results_table_sepsis}. As the sepsis prediction task has a very strong class imbalance, we introduce the F1 score as an additional metric. For the sake of completeness, we also report accuracy and AUC such as for the other experiments, but due to the high imbalance, these metrics are less meaningful. The performance development (F1 score) in relation to the label ratio is visualized in figure \ref{development_sepsis}. In comparison to the self-supervised settings, where the pre-training effect decreases for higher label ratios, the improvement in transfer learning remains nearly constant for different label ratios. This indicates as we pre-train on a different dataset than we fine-tune, the size difference between the pre-training and fine-tuning set is less relevant. Further, when comparing the from scratch performance to the performance of the pre-trained model, it can be seen that the AUC improves in most cases and the F1 score improves for all label ratios, while accuracy is mostly similar for both models. This shows, that the pre-trained model handles the class imbalance better and improves in predicting the rare positive class by reducing the false negative predictions. This is an important property, as for sepsis prediction it is especially important to classify the positive cases correctly to intervene in time. Overall, even though the MIMIC-III and the Sepsis Prediction dataset encompass different features, we show that pre-training on MIMIC-III can be used to improve sepsis prediction performance notably, implying that the knowledge acquired during pre-training supports a general understanding of clinical population graphs and is not dataset-specific. This broadens the use case of our method substantially, as it allows pre-training on data with differing features. Thereby, for example, a small hospital with limited access to (labeled) data can use data from larger hospitals for pre-training. Further, it also allows to pre-train with data that was recorded with another application in mind and thus does not exactly match the features of the task at hand.

\begin{table}[htb!]
    \caption{Performance of the proposed model in accuracy, AUC, and F1 score trained from scratch (SC) or fine-tuned after pre-training on MIMIC-III (FT) for different label ratios on the Sepsis prediction task.}
  \centering
    \begin{tabular}{cccc}
        \toprule
         \multicolumn{4}{c}{SEPSIS} \\
        \midrule
        Labels & Metric & SC & FT\\
        \midrule
        1\% & ACC & \textbf{89.32 $\pm$ 1.23} & 88.95 $\pm$ 2.01\\
        & AUC & 68.18 $\pm$ 4.29 & \textbf{70.25 $\pm$ 3.01}\\
        & F1 & 56.32 $\pm$ 1.44 & \textbf{58.28 $\pm$ 1.40}\\
        \midrule
        5\% & ACC & \textbf{89.91 $\pm$ 1.12} & 89.84 $\pm$ 0.49\\
        & AUC & 75.60 $\pm$ 4.47 & \textbf{77.16 $\pm$ 2.05}\\
        & F1 & 60.74 $\pm$ 1.31 & \textbf{61.65 $\pm$ 1.77}\\
        \midrule
        10\% & ACC & \textbf{91.12 $\pm$ 0.45} & 90.86 $\pm$ 1.16\\
        & AUC & \textbf{80.04 $\pm$ 0.86} & 79.80 $\pm$ 1.26\\
        & F1 & 62.70 $\pm$ 1.98 & \textbf{64.43 $\pm$ 2.35}\\
        \midrule
        50\% & ACC & 91.32 $\pm$ 0.71 & \textbf{92.47 $\pm$ 1.15}\\
        & AUC & 80.98 $\pm$ 1.77 & \textbf{83.73 $\pm$ 0.93}\\
        & F1 & 63.95 $\pm$ 2.13 & \textbf{66.94 $\pm$ 1.33}\\
        \midrule
        100\% & ACC & 92.49 $\pm$ 0.38 & \textbf{92.76 $\pm$ 0.31}\\
        & AUC & 83.40 $\pm$ 1.38 & \textbf{84.49 $\pm$ 1.27}\\
        & F1 & 64.62 $\pm$ 1.05 & \textbf{67.59 $\pm$ 1.31}\\
        \bottomrule
    \end{tabular}
    \label{results_table_sepsis}
\end{table}

\subsubsection{Effect of Embedding Space}

\begin{figure*}[htb!]
\centering
\includegraphics[width=\linewidth]{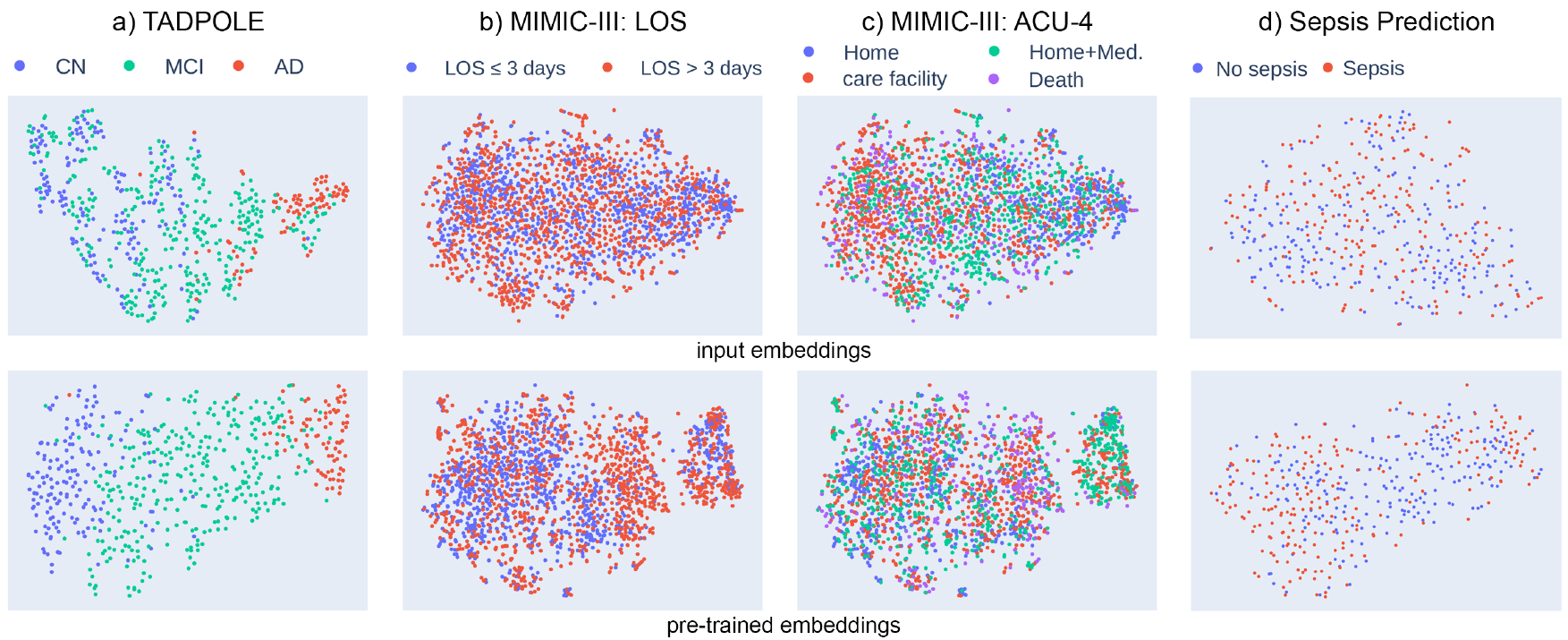}
\caption{T-sne visualizations of input and pre-trained embeddings for disease prediction on TADPOLE, LOS, and ACU-4 prediction on MIMIC-III and early prediction of sepsis. The upper row shows the input embeddings and the lower row the pre-trained embeddings. For MIMIC-III we only used the embeddings of five graphs to keep the visualization clear. For the Sepsis Prediction dataset, the model was pre-trained on MIMIC-III with multi-task pre-training.}
\label{tsne}
\end{figure*}

To give an intuition of why the model initialization learned during pre-training is superior to a random model initialization, we show the effect of pre-training on the distribution of the patient record embeddings (Figure \ref{tsne}). We compare the embeddings computed by the pre-trained model to the embeddings computed by a randomly initialized model. To visualize these embeddings we use t-sne for dimensionality reduction and visualize the data in a 2D space. Figure \ref{tsne} shows the visualizations for both the untrained and the pre-trained models, which are extracted directly before the decoder layer. We used Feature Masking as a pre-training task for these visualizations.

Even though no labels were used for pre-training, the classes are clearly better separated for the pre-trained embeddings. Especially for the TADPOLE dataset, the classes are quite well separated after pre-training (Figure \ref{tsne} a)). This explains why for small label ratios, where from scratch training achieves bad results, the effect of pre-training is very high. For the MIMIC-III dataset, pre-training does not lead to a clear class separation, however, in the pre-trained embeddings more clusters of single classes can be seen (Figure \ref{tsne} b),c)). A similar effect can be observed for the Sepsis Prediction dataset. Even though pre-training was performed on MIMIC-III, the septic patients seem to be clustered at the sides of the space, while the non-septic patients are more frequent in the center of the space (Figure \ref{tsne} d)). When fine-tuning given this pre-trained initialization, which already clusters nodes with the same class together, the model can easier learn to separate the classes, enabling it to quickly learn a good separation even with little labeled data.

\subsubsection{Ablation Studies}
We perform several ablation studies to evaluate different parts of our proposed model on pre- and downstream task training. All pre-training ablations are performed with limited labels to ensure a well-observable effect in fine-tuning. Specifically, we use a label ratio of 10\% on the larger MIMIC-III dataset and of 1\% on the TADPOLE dataset. Further, for MIIMC-III we selected one pre-training method and one downstream task for the ablation studies. We perform all ablations using the LOS task and Feature Masking as pre-training, as this pre-training method was overall the best performing one. Further, we analyze the contribution of the different pre-training tasks during Multi-Task Pre-training. We further performed an analysis of the wrongly classified samples, which can be found in the supplementary material.

\textbf{Effect of Graphormer:} We replace the Graphormer module with a linear layer or different graph neural networks (GCN (\cite{gcn_kipf}), GAT (\cite{velickovic2018graph}) and GIN (\cite{xu2018powerful})) and train the model from scratch on the full dataset (see results in Table \ref{ablations_mimic} a) and Table \ref{ablations_tadpole} a)). The number of layers, the learning rate, and the training epochs are optimized for every conducted ablation.

On the MIMIC-III dataset, the proposed model outperforms the linear model as well as all models based on other GNN architectures, showing that for this more complex dataset using a population graph as well as the use of the attention-based Graphormer layers are beneficial. For TADPOLE, the linear model reaches slightly better performance in terms of AUC, while the less complex GNNs reach inferior performance compared to the more complex Graphormer model. The good performance of the linear model shows, that using a population graph is not as beneficial for TADPOLE, which is a relatively small and easy dataset. Instead, each node's features give sufficient information to solve the downstream task. This is also apparent when considering the worse performance of GCN, GAT, and GIN. These models are based on neighborhood aggregation and in each update step the information of only the neighbors can be used, which seems to have a negative effect. Graphormer on the other hand can counteract this effect by using a node-level attention mechanism over the full graph, making it possible to select information from specific nodes in the graph or none at all.

Further, we perform pre-training followed by fine-tuning with limited labels for the different ablations of our model and compare the effect of pre-training the different ablation models with the effect on our proposed model. These results are shown in Table \ref{ablations_mimic} b) and Table \ref{ablations_tadpole} b). Our proposed unsupervised pre-training method proves to be beneficial for the linear model as well as for the other GNN models, indicating that the pre-training technique is transferable to other models as well. However, the effect of pre-training is highest for the proposed graph transformer based model, showing the value of combining transformer based models with masking-based pre-training tasks. On MIMIC-III we can also see a benefit of using a graph model compared to the linear model in the pre-training effect, showing that for this complex dataset the modeling and processing as a population graph during pre-training has a positive effect.

\begin{table*}[htb!]
\centering 
\caption{Ablations to test the Graphormer module by replacing it with a linear layer or different GNNs on MIMIC-III a) downstream task performance trained from scratch b) results of fine-tuning (FT) on 1\% labels, compared to training from scratch (SC) c) performance in pre-training task.}
\begin{tabular}{ccccccc}
\toprule
 &      & Linear                    & GCN                   & GAT              & GIN                     & Proposed                   \\\midrule
\multirow{2}{*}{a}  & ACC  & 67.25 $\pm$ 01.11      & 68.74 $\pm$ 01.50     & 69.22 $\pm$ 1.35 & 68.94 $\pm$ 1.11 & \textbf{70.29 $\pm$ 01.10} \\
  & AUC  & 72.69 $\pm$ 00.97      & 72.64 $\pm$ 01.10     & 74.43 $\pm$ 1.10 & 74.84 $\pm$ 1.18 &  \textbf{76.17 $\pm$ 01.02} \\\midrule
\multirow{4}{*}{b} & ACC  & 64.71 $\pm$ 0.84           & 61.77 $\pm$ 2.22          & 62.79 $\pm$ 0.91     & 61.78 $\pm$ 1.37     & \textbf{69.42 $\pm$ 1.23}      \\
  & gain & +0.93                  & +1.73                 & +2.34            & +1.10            & \textbf{+4.70}             \\
  & AUC  & 67.94 $\pm$ 1.20           & 66.03 $\pm$ 2.40          & 66.57 $\pm$ 2.03     & 65.65 $\pm$ 1.77     &  \textbf{75.09 $\pm$ 1.29}               \\
  & gain & +0.22                  & +3.09                 & +3.11            & +2.13            &  \textbf{+6.12 }                     \\\midrule
\multirow{2}{*}{c} & RMSE & 0.838 $\pm$ 0.015& 0.941 $\pm$ 0.027         & 0.810 $\pm$ 0.352    & 0.951 $\pm$ 0.029    &  \textbf{0.783 $\pm$ 0.011}     \\
  & F1  & 77.59 $\pm$ 0.61           & 69.97 $\pm$ 0.39 & 81.35 $\pm$ 0.51     & 70.64 $\pm$ 0.59     & \textbf{81.58 $\pm$ 0.41}             \\   
  \bottomrule 
\end{tabular}
\label{ablations_mimic}
\end{table*}

\begin{table*}[htb!]
\centering 
\caption{Ablations to test the Graphormer module by replacing it with a linear layer or different GNNs on TADPOLE a) downstream task performance trained from scratch b) results of fine-tuning (FT) on 1\% labels, and performance gain compared to training from scratch c) performance in pre-training task.}
\begin{tabular}{ccccccc}
\toprule
 &      & Linear               & GCN               & GAT              & GIN                    & Proposed                   \\\midrule
\multirow{2}{*}{a}  & ACC  & 91.14 $\pm$ 02.62 & 74.27 $\pm$ 06.41 & 73.09 $\pm$ 5.94 & 73.47 $\pm$ 6.85  & \textbf{92.59 $\pm$ 03.64} \\
  & AUC  & \textbf{97.77 $\pm$ 01.59} & 89.89 $\pm$ 04.12 & 89.65 $\pm$ 4.39 & 89.40 $\pm$ 4.81 &  96.96 $\pm$ 02.23 \\\midrule
\multirow{4}{*}{b} & ACC  & 73.56 $\pm$ 12.30      & 61.87 $\pm$ 5.03      & 60.58 $\pm$ 5.46     &  59.59 $\pm$ 4.80     & \textbf{74.39 $\pm$ 7.58}      \\
  & gain & +12.26            & +3.67             & +0.45            & +5.02            &  \textbf{+14.97}            \\
  & AUC  & \textbf{92.33 $\pm$ 3.47}      & 77.68 $\pm$ 4.84      & 74.52 $\pm$ 6.08     & 77.25 $\pm$ 5.05     &  89.98 $\pm$ 3.57               \\
  & gain & +18.33             & +9.62             & +3.14            & +9.17            &  \textbf{+21.26}                     \\\midrule
\multirow{2}{*}{c} & RMSE & 0.140 $\pm$ 0.010     & 0.140 $\pm$ 0.020     & 0.131 $\pm$ 0.018    & 0.126 $\pm$ 0.012    &  \textbf{0.116 $\pm$ 0.008}              \\
  & ACC  & 65.33 $\pm$ 4.76      & \textbf{81.71 $\pm$ 5.25}      & 80.78 $\pm$ 3.67     & 80.86 $\pm$ 5.12     &  69.47 $\pm$ 6.20    \\   
  \bottomrule
\end{tabular}
\label{ablations_tadpole}
\end{table*}

Table \ref{ablations_mimic} c) and Table \ref{ablations_tadpole} c) show the masked imputation performance during pre-training, measured by RMSE for continuous features, so measurement features for MIMIC-III and imaging features for TADPOLE, F1 score for the discrete treatment features in MIMIC-III and accuracy for the discrete TADPOLE features (apoe4+cognitive tests). For the cognitive tests in TADPOLE, we use clinically-motivated error margins in which a prediction is considered correct, to compute the accuracy. The used margins are shown in Table \ref{error_margins}. The predicted cognitive test scores are ordinal features, so close scores imply a similar performance of the patient. As many possible scores exist, it is a very hard task to make an accurate prediction. By applying error margins, we can assess better, if the model predicts test results in the correct range of values. Overall the results show, that the performance in the pre-training task is not always directly linked to the benefit of pre-training. Especially on TADPOLE, it can be observed, that while GCN, GAT, and GIN perform very well in predicting the cognitive test results, their pre-training benefit is restricted. However, their performance in predicting the imaging features is less than for the proposed model. On MIMIC-III, the proposed Graphormer-based model outperforms all other models in the pre-training task, which is in line with the superior fine-tuning results. In summary, while Graphormer has not always the best performance in the pre-training task, the extracted knowledge is used more beneficially for improving the downstream task, leading to a better fine-tuning performance on both datasets.
\begin{table}[htb!]
  \caption{Clinically-motivated, feature-dependent error margins for measuring cognitive test results prediction performance during pre-training on TADPOLE.}
  \centering
    \begin{tabular}{p{3cm}cc}
        \toprule
        cognitive test & number of classes & error margin\\
        \midrule
        CDRSB & 19 & 4\\
        ADAS11 & 107 & 15\\
        MMSE & 11 & 2\\
        RAVLT\_immediate & 68 & 5 \\
        \bottomrule
    \end{tabular}
    \label{error_margins}
\end{table}

\begin{figure}[htb!]
\centering
\includegraphics[width=\linewidth]{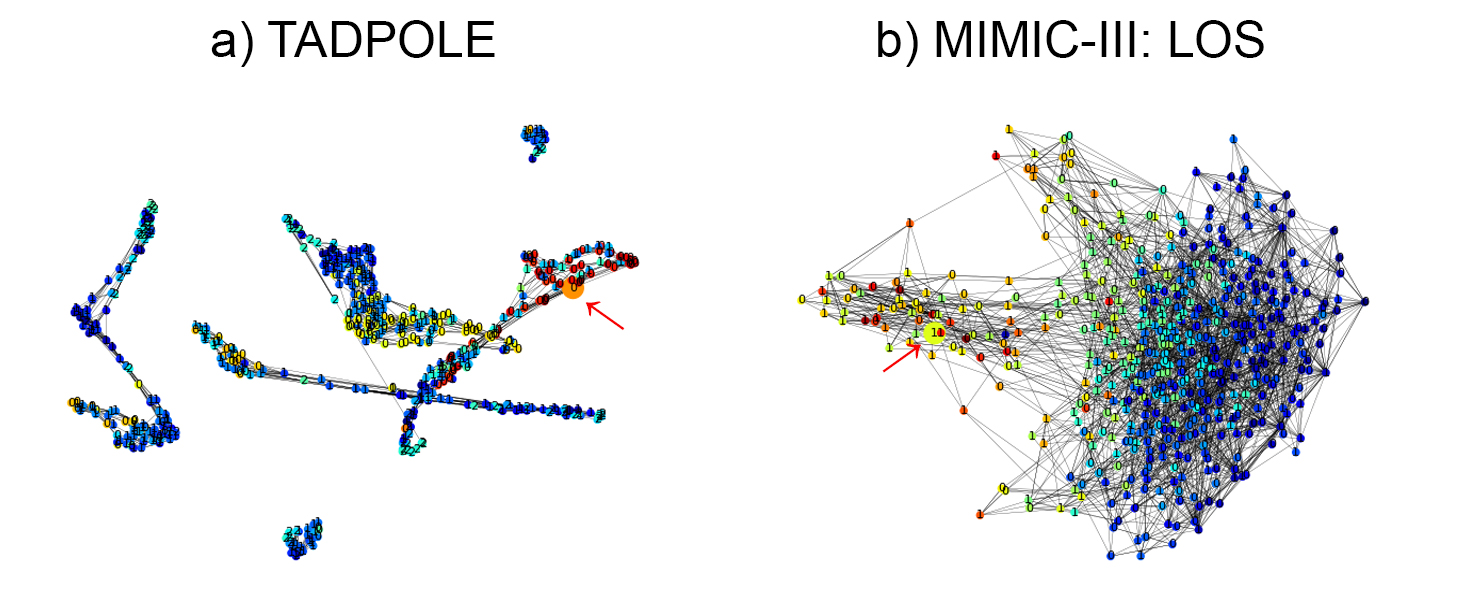}
\caption{Visualization of the node-level attention of Graphormer. The used color map displays the highest attention in red, going to yellow, green, and blue for the lowest attention. The current node for which the attention is computed is visualized larger in the image and indicated by the red arrows. The attention is distributed over the whole graph but focuses more on closer nodes.}
\label{attention_visualization}
\end{figure}

Moreover, to better understand the effect of the patient-level attention in the graph transformer module, we visualize the learned attention of the best-performing models for two samples from the TADPOLE and MIMIC-III dataset for Alzheimer’s disease and length-of-stay prediction in Fig. \ref{attention_visualization}. We compute the attention by averaging over the layers and heads of the model. It can be observed that the attention to patients close to the current patient is higher than to more distant ones. Moreover, it can be seen that while the current patient is always attended notably, the model also takes other patients into account, showing the model makes use of the information provided in the graph.

\textbf{Effect of Transformer:}
For MIMIC-III, we insert transformer layers in the encoder to deal with time-series data. To test the effect of these transformer layers, we compare the performance with and without these layers for both from scratch training and fine-tuning. The results of this ablation are shown in Table \ref{ablation_transformer}. Part a) in Table \ref{ablation_transformer} shows the results, when training the proposed model from scratch on the full dataset with and without the transformer layer. This results in a reduction of accuracy and AUC, showing that the transformer layers are helpful for processing the time-series input data of MIMIC-III. 

For pre-training on MIMIC-III, the model needs to predict time-dependent outputs, suggesting that here the transformer layers should be especially important, as they can understand the temporal context. To investigate the effect of transformer during pre-training, we remove the transformer layer and compare the effect of pre-training the model when fine-tuning it with limited labels, with and without the transformer layer (see Table \ref{ablation_transformer} b)). Further, we show the performance in the pre-training task for both models (see Table \ref{ablation_transformer} c)). For training from scratch with limited labels, the use of transformer layers improves performance. However, the fine-tuned model without transformer layers reaches a slightly better accuracy and a slightly worse AUC than with the transformer layer. Overall the performance is very similar to the fine-tuned model with transformer layers. Likewise, the performance in the pre-training task of these two models is very similar. This indicates that the pre-trained Graphormer-based model is expressive enough to learn and benefit from the pre-training task and thus less reliant on the transformer layers.

Overall, the transformer layers improve model performance when trained from scratch. However, for pre-training Graphormer, they have little effect, as Graphormer alone already performs well in the pre-training task and benefits from it in fine-tuning.

\begin{table}[hbt!]
  \caption{All results when removing the transformer layers on MIMIC-III. a) proposed model on the full dataset without pre-training b) from scratch training and fine-tuning with 10\% labels c) performance in pre-training task.}
  \centering
\begin{tabular}{cccc}
\toprule
  &      & without transformer   & with transformer      \\\midrule
\multirow{2}{*}{a} & ACC  & 69.39 $\pm$ 0.60          & \textbf{70.29 $\pm$ 1.10 }         \\
  & AUC  & 75.03 $\pm$ 1.23          & \textbf{76.17 $\pm$ 1.02}          \\\midrule
\multirow{6}{*}{b} & ACC - SC  & 63.79 $\pm$ 0.54          & \textbf{64.72 $\pm$ 0.45}          \\
  & ACC - PT  & 69.34 $\pm$ 1.05          & \textbf{69.42 $\pm$ 1.23}          \\
  & gain & \textbf{+5.55}        & +4.70                 \\\cline{2-4}
  & AUC - SC  & 68.41 $\pm$ 1.51          & \textbf{68.97 $\pm$ 0.66}          \\
  & AUC - PT  & \textbf{75.19 $\pm$ 1.03}          & 75.09 $\pm$ 1.29          \\
  & gain & \textbf{+6.78}        & +6.12                 \\\midrule
\multirow{2}{*}{c} & RMSE & \textbf{0.770 $\pm$ 0.016} & 0.783 $\pm$ 0.011         \\
  & F1  & 80.06 $\pm$ 0.49          & \textbf{81.58 $\pm$ 0.41} \\\bottomrule
\end{tabular}
\label{ablation_transformer}
\end{table}

\textbf{Multi-task Pre-Training:}
Our results show that combining different pre-training tasks can help to improve downstream prediction results after fine-tuning even further. In our experiments, we combined all four pre-training tasks we proposed for clinical time-series data as given in the MIMIC-III dataset. To analyze how important each of the four proposed tasks is for multi-task pre-training, we pre-train our model once with all combinations of three pre-training tasks, always leaving out the fourth task. We then fine-tune the models for Length-of-Stay prediction on MIMIC-III and compare the results as shown in Table \ref{ablation_table_MT}. Our main observation is, that using all four tasks in most cases delivers the best results or is very close to the best result after fine-tuning on the downstream task. This indicates that all four tasks are important to improve the model's understanding of some aspects of the data. Further, leaving our Patient or Feature Masking leads to the overall worst results, while when leaving out Treatment Prediction or Block-wise Masking the performance remains higher. This shows that Patient and Feature Masking have a higher contribution to the overall performance. This aligns with the results from pre-training with single pre-training tasks for LOS, as here these two tasks also lead to the highest benefit for downstream task performance.

\begin{table*}[htb!]
\centering
\caption{Results of multi-task pre-training with all four tasks compared to training with all but one task. The best result per label ratio is bold, while the worst and second-worst results are marked with bold red and red. Fine-tuning on LOS prediction. 
}
\begin{tabular}{ccccccc}
\toprule
\multicolumn{7}{c}{MIMIC-III: Multitask Pre-training}                                                                                                                          \\ \midrule
Labels  & Metric & MT & MT: no TP        & MT: no BM                                       & MT: no TFM    &    MT: no PM         \\ \midrule
1\%   & ACC    &  \textbf{66.39 $\pm$ 1.34}& \color{red}\textbf{66.01 $\pm$  0.77} & \color{red}66.07 $\pm$ 1.74 & 66.37 $\pm$ 1.37 & 66.11 $\pm$ 1.12    \\
      & AUC    &  71.40 $\pm$ 1.81& \color{red}\textbf{70.57 $\pm$ 1.26} & \color{red}71.08 $\pm$ 2.36  & \textbf{71.45 $\pm$ 1.46}&  71.42 $\pm$ 1.76   \\\midrule
5\%   & ACC    &  68.97 $\pm$ 0.46& 68.98 $\pm$ 0.98 & \textbf{69.15 $\pm$ 1.07}  & \color{red}\textbf{68.59 $\pm$ 0.56}  &  \color{red}68.71 $\pm$ 1.06 \\
      & AUC    &  \textbf{74.99 $\pm$ 1.00}& 74.81 $\pm$ 1.08 & \color{red}74.62 $\pm$ 1.36  & \color{red}\textbf{74.32 $\pm$ 1.05} &  74.78 $\pm$ 1.21 \\ \midrule
10\%  & ACC    &  \textbf{69.99 $\pm$ 0.97}& 69.54 $\pm$ 1.05 & 69.66 $\pm$ 0.60 & \color{red}69.10 $\pm$ 0.71 &  \color{red}\textbf{69.03 $\pm$ 1.21}          \\
      & AUC    &  \textbf{75.90 $\pm$ 1.35}& 75.53 $\pm$ 1.06 & 75.70 $\pm$ 1.10 & \color{red}\textbf{74.77 $\pm$ 0.48} &  \color{red}75.40 $\pm$ 1.09   \\ \midrule
50\%  & ACC    &  \textbf{71.46 $\pm$ 0.91}& 71.10 $\pm$ 0.56 & 71.29 $\pm$ 1.12 & \color{red}\textbf{70.85 $\pm$ 0.88}  &  \color{red}71.01 $\pm$ 0.92   \\
      & AUC    &  77.77 $\pm$ 1.17& \textbf{77.79 $\pm$ 0.85} & 77.58 $\pm$ 1.23  & \color{red}77.23 $\pm$ 1.15 &  \color{red}\textbf{76.98 $\pm$ 0.91}\\ \midrule
100\% & ACC    &  \textbf{72.13 $\pm$ 1.46}& 72.03 $\pm$ 0.61 & \color{red}71.53 $\pm$ 0.62  & \color{red}\textbf{71.34 $\pm$ 0.85} &  71.61 $\pm$ 0.64    \\
      & AUC    &  \textbf{78.73 $\pm$ 1.21} & 78.69 $\pm$ 0.91 & 78.44 $\pm$ 1.22 & \color{red}78.27 $\pm$ 1.00 &  \color{red}\textbf{77.88 $\pm$ 0.78}\\ \bottomrule
\end{tabular}
\label{ablation_table_MT}
\end{table*}

\textbf{Effect of high-dimensional imaging data:}

We demonstrated the applicability of our method for fusing imaging and non-imaging data on the TADPOLE dataset, which combines features extracted from MRI and PET with cognitive test results, demographic data and, genetics. To further analyze the capability of our model when using high-dimensional imaging features and showing the promise of fusing multi-modal features we extended our experiments on TADPOLE to a higher number of input features. The increased feature set includes 9 cognitive test results, 7 genetic and demographic features as well as 326 imaging features from three imaging modalities (MRI, PET, DTI), including various biomarkers and some image quality measures. Further, we analyze the performance of our model when restricting the input to only imaging and only non-imaging features. Table \ref{tadpole_full_feats} shows the results of this experiment. It can be observed that the performance of our model remains very similar when introducing the high-dimensional imaging features, showing the robustness of our model to this kind of data. When using only imaging features the performance drops significantly, and also dropping the imaging features leads to a performance decrease. This shows the importance of the multi-modal feature fusion our method enables.

\begin{table}[htb!]
  \caption{Results on TADPOLE dataset with an increased feature set. Further, we analyze the model performance when restricting the input to only imaging or non-imaging features.}
  \small
  \centering
    \begin{tabular}{cccccc}
        \toprule
        & ours & all features & imaging & non-imaging \\
        \midrule
        AUC & 96.96$\pm$2.13 & 96.14$\pm$3.37 & 72.74$\pm$2.18 & 95.53$\pm$2.33 \\
        \midrule
        ACC & 92.59$\pm$3.64 & 91.31$\pm$4.85 & 42.90$\pm$2.90 & 89.39$\pm$2.56 \\
        \bottomrule
    \end{tabular}
    \label{tadpole_full_feats}
\end{table}

Furthermore, our general approach can be not only valuable in the analysis of static imaging biomarkers such as in TADPOLE, but also for temporal imaging biomarkers, extracted from dynamic imaging modalities such as dynamic MRI (\cite{potsch2021ai, zheng2021imaging}), PET (\cite{noortman2020adding}) or Ultrasound (\cite{ma2021dynamic}). Further, it can be applied to longitudinal imaging studies with frequent measurements, such as in the e.g. positioning verification of non-small cell lung cancer, where CT scans are usually acquired daily and their features could be used for treatment adaption (\cite{van2017feature}).

We further demonstrate that our model can be adapted to a setting including spatial images. Following \cite{soenksen2022integrated}, who analyzed the use of multi-modal data for patient outcome prediction, we collect fused data from the MIMIC-IV (\cite{johnson2020mimic}) and the MIMIC-CXR (\cite{johnson2019mimic_cxr}) datasets. Each patient record includes demographic, measurement, and treatment data very similar as described in section 4.1.1. (MIMIC-III). Further, each record includes one Chest X-Ray image of the patient. We use this dataset in order to test the applicability of our model in a setup where numerical EHR information, as well as spatial imaging data, is available. We consider the task of 48h mortality prediction given the demographics of the patient, an X-Ray image at a time-point t, and measurement and treatment data about the last 24 hours before time-point t. The task is to predict whether the patient will die within the next 48 hours. To integrate the spatial imaging information into our model, we employ a chest X-Ray encoder based on a DenseNet backbone, pre-trained for ChestXRay pathology classification \cite{cohen2022torchxrayvision}. The used image encoder has around 6.9 million parameters, of which we fine-tune the last block of 2.2 million parameters. In contrast, the graph encoder has around 5.6 million parameters. Our experiments showed that these encoder sizes were a good trade-off avoiding overfitting but allowing to capture all information in the data. We apply a late fusion approach to fuse the embedding representation given by the X-ray encoder with the output of our data embedding module. For this, we concatenate the two representations before processing them with our decoder.

Table \ref{imaging_results} shows the results of this experiment. We observe a substantial performance increase by fusing the imaging and non-imaging data compared to only considering a single data source, demonstrating that our model successfully fuses the multi-modal data and showing the applicability of our model for tasks including spacial imaging data.  In the context of pre-training, integrating imaging data needs further investigation, as our pre-training method is primarily designed for understanding heterogeneous but numerical inputs. Therefore, our pretrained EHR encoder may not be sufficient for effectively capturing the relationships between the EHR data and the imaging data. Further, during the pre-training phase, our method did not encounter any spatial imaging data or the combination of EHR and imaging data. As a result, the pre-trained model may not have developed an adequate understanding of the relationships between these two data types, which could have contributed to the lack of improvement in the new setting. We believe it is an important step for applications involving spatial image information to investigate joined pre-training on spacial imaging data and other EHR data, which could be a potential future work.

Overall, our results show the applicability of our model to settings with spatial imaging data and further highlight the importance of considering comprehensive patient data in the field of medical imaging analysis.

\begin{table}[htb!]
  \caption{Results on the merged MIMIC-IV and MIMIC-CXR dataset for mortality prediction using only imaging, only non-imaging information or fused information. We further show the effect of our pre-training in this setting.}
  \small
  \centering
    \begin{tabular}{ccccc}
    \hline & imaging & non-imaging & fused & fused PT \\
    \hline AUC  & 73.10$\pm$6.91 & 82.95$\pm$1.37 & \textbf{85.87$\pm$3.37}& 84.50$\pm$1.74 \\
    \hline ACC & 67.90$\pm$9.82 & 74.09$\pm$4.73 & \textbf{78.54$\pm$3.79} & 75.24$\pm$2.99 \\
    \hline
    \end{tabular}
    \label{imaging_results}
\end{table}

%% file: chapters/conclusion.tex
\section{Conclusion} \label{conclusion}
In this paper, we propose multiple novel unsupervised pre-training methods for multi-modal clinical record data based on masked imputation. We propose to model the clinical data in a patient population graph, such that the model can use other patients' information for both pre-training and fine-tuning. For this setup, we present several pre-training tasks, designed to learn a general understanding of multi-modal clinical data. Moreover, we propose a network architecture based on transformer and graph transformer for pre-training and learning on patient population graphs built from heterogeneous clinical data. We show the superiority of the proposed pipeline for various prediction tasks on three datasets and provide an extensive analysis of our method showing different ablation studies. We test our method on MIMIC-III and TADPOLE for self-supervised pre-training and a Sepsis Prediction dataset in a transfer learning setup. All datasets contain medical patient records but encompass different features. We show that pre-training improves results for all datasets over the full dataset size both in the self-supervised as well as the transfer learning scenario. Overall, pre-training proves to be especially helpful for scenarios where only a limited amount of labeled data is used for fine-tuning. Moreover, we compare our different pre-training approaches to one another, showing that more complex tasks tend to have a larger positive impact on the prediction results of downstream tasks. Finally, we combine our proposed pre-training tasks in a multi-task setup, leading to an additional performance gain. Our pre-training methods are all unsupervised and as such task-independent, making them likewise applicable to self-supervised pre-training and transfer learning. Thus the proposed pipeline opens a path to improve learning over multi-modal clinical data on small datasets or datasets with limited labels via pre-training on large unlabeled collections of clinical records from either the same or a different data source.\\

\noindent\textbf{Acknowledgements}\\
Anees Kazi's financial support was provided by BigPicture (IMI945358). Data collection and sharing for one of the datasets used this project was funded by the Alzheimer's Disease Neuroimaging Initiative (ADNI) (National Institutes of Health Grant U01 AG024904) and DOD ADNI (Department of Defense award number W81XWH-12-2-0012). ADNI is funded by the National Institute on Aging, the National Institute of Biomedical Imaging and Bioengineering, and through generous contributions from the following: AbbVie, Alzheimer’s Association; Alzheimer’s Drug Discovery Foundation; Araclon Biotech; BioClinica, Inc.; Biogen; Bristol-Myers Squibb Company; CereSpir, Inc.; Cogstate; Eisai Inc.; Elan Pharmaceuticals, Inc.; Eli Lilly and Company; EuroImmun; F. Hoffmann-La Roche Ltd and its affiliated company Genentech, Inc.; Fujirebio; GE Healthcare; IXICO Ltd.; Janssen Alzheimer Immunotherapy Research \& Development, LLC.; Johnson \& Johnson Pharmaceutical Research \& Development LLC.; Lumosity; Lundbeck; Merck \& Co., Inc.; Meso Scale Diagnostics, LLC.; NeuroRx Research; Neurotrack Technologies; Novartis Pharmaceuticals Corporation; Pfizer Inc.; Piramal Imaging; Servier; Takeda Pharmaceutical Company; and Transition Therapeutics. The Canadian Institutes of Health Research is providing funds to support ADNI clinical sites in Canada. Private sector contributions are facilitated by the Foundation for the National Institutes of Health (www.fnih.org). The grantee organization is the Northern California Institute for Research and Education, and the study is coordinated by the Alzheimer’s Therapeutic Research Institute at the University of Southern California. ADNI data are disseminated by the Laboratory for Neuro Imaging at the University of Southern California.

\FloatBarrier